\documentclass[11pt]{article}

\usepackage[preprint]{acl}

\usepackage{times}
\usepackage{latexsym}

\usepackage[T1]{fontenc}
\usepackage[utf8]{inputenc}

\usepackage{microtype}

\usepackage{inconsolata}

\usepackage{graphicx}

\usepackage{hyperref}       % hyperlinks
\usepackage{url}            % simple URL typesetting
\usepackage{booktabs}       % professional-quality tables
\usepackage{amsfonts}       % blackboard math symbols
\usepackage{nicefrac}       % compact symbols for 1/2, etc.
\usepackage{microtype}      % microtypography
\usepackage{xcolor}         % colors
\usepackage{pifont}
\usepackage{amsmath}
\usepackage{tcolorbox}
\usepackage{enumitem}
\usepackage{multirow}
\usepackage{multicol}
\DeclareMathOperator*{\argmax}{arg\,max}
\usepackage{tabularx}

\title{Diagnosing the Explicit-to-Applied Theory of Mind Gap in LLMs}

\author{
 \textbf{Prajvi Saxena\textsuperscript{1}},
 \textbf{Sabine Janzen\textsuperscript{1}},
 \textbf{Wolfgang Maaß\textsuperscript{1,2}},
\\
 \textsuperscript{1}German Research Center for Artificial Intelligence (DFKI),
 \textsuperscript{2}Saarland University
\\
 \small{
   \textbf{Correspondence:} \href{mailto:email@domain}{prajvi.saxena@dfki.de}
 }
}

\begin{document}
\maketitle
\begin{abstract}
As large language models are deployed in high-stakes roles such as therapy, education, and personal assistance, their social cognition must be assessed reliably. Current Theory of Mind (ToM) benchmarks report aggregate scores that conflate two distinct competencies: explicit ToM (recognizing what someone believes) and applied ToM (using that belief to predict behavior). Building on this distinction, we investigate the gap and its recoverability using four structurally diverse benchmarks. Aggregate reporting obscures uneven competence: on SimpleToM, Llama-3.1-8B scores 51.5\% overall while recognizing mental states at 99.4\% and predicting behavior at only 28.9\%. To assess recovery, we compare structured prompting, matched in-context learning, and our proposed \textbf{Belief-State Test-Time Training (BSTTT)}, which adapts lightweight parameters on answered support questions through action reconstruction. Across Qwen2.5-7B, Llama-3.1-8B, and GPT-oss-20B, recovery depends on the model, task, and intervention. On SimpleToM, reminders combined with BSTTT improve Llama’s behavior prediction by 40.8 percentage points; reminders account for most of the overall improvement, with BSTTT adding 1.3–5.5 points beyond reminders across models. Contextual examples often outperform weight updates, while BSTTT remains stronger in specific model–benchmark settings. Temporal belief tracking remains difficult despite partial recovery. These findings show that aggregate ToM scores can overstate applied competence while hiding performance recoverable through targeted assistance. We recommend decomposed reporting of explicit and applied accuracy alongside intervention comparisons that establish which failures are recoverable and under what conditions. Our code repository is available at \footnote{GitHub: \url{https://anonymous.4open.science/r/TestTimeTraining-of-LLMs-E67F/README.md}}.
\end{abstract}

\section{Introduction}
\label{sec:intro}

\begin{figure*}[t]
    \centering
    \small
    \includegraphics[width=\textwidth]{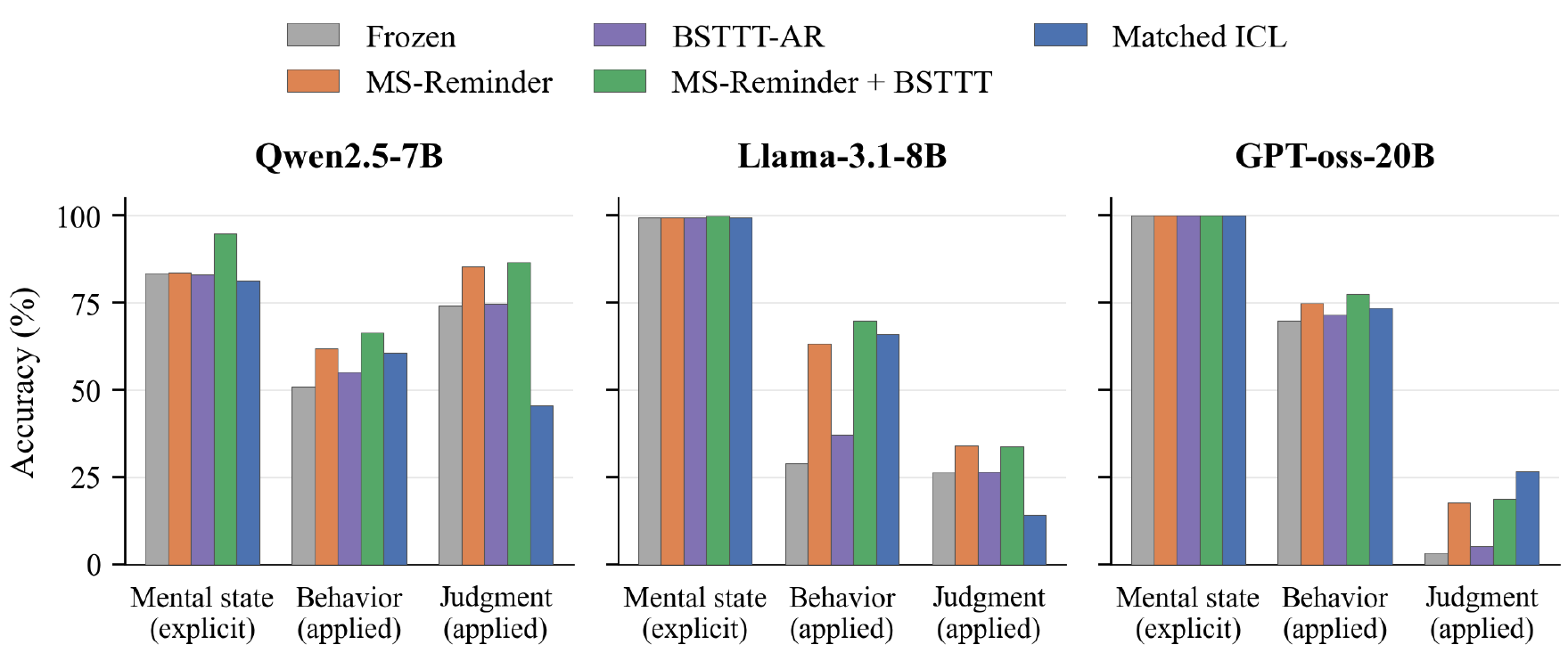}
    \caption{\textbf{The explicit-to-applied Theory of Mind gap on SimpleToM} Each scenario generates three matched questions: mental-state recognition (\textit{explicit}), behavior prediction (\textit{applied}), and judgment (\textit{applied}).}
    \label{fig:gap}
\end{figure*}

Theory of Mind (ToM) is the capacity to attribute beliefs, intentions, and emotions to other agents and to reason about how those states shape behavior~\citep{Premack_Woodruff_1978, BARONCOHEN198537}. This capacity encompasses two qualitatively different competencies. \textbf{Explicit ToM} is recognizing that another agent holds a particular mental state. \textbf{Applied ToM} is using that recognition to predict behavior, anticipate consequences, or evaluate actions. The two can dissociate~\citep{apperly2012theory}. Reporting a single accuracy that averages recognition and application questions obscures this distinction. Figure~\ref{fig:gap} illustrates the dissociation on SimpleToM: Llama-3.1-8B recognizes mental states at 99.4\% yet predicts the resulting behavior at only 28.9\%, a 70.5-point explicit-to-applied gap. Its overall accuracy of 51.5\% hides both near-perfect recognition and weak application.\\
This explicit-to-applied gap motivates two hypotheses. Under a \textbf{capacity-limitation hypothesis (H1)}, models lack some of the reasoning capabilities required to translate recognized mental states into reliable behavioral predictions. Under a \textbf{latent-competence hypothesis (H2)}, models possess relevant social knowledge but fail to consistently apply it under standard inference. These hypotheses motivate different remedies: improving the underlying capabilities or improving how existing capabilities are elicited.\\
Prompt-level interventions, including mental-state reminders, perspective taking~\citep{wilf2024think}, and symbolic belief tracking~\citep{sclar-etal-2023-minding}, test recoverability without parameter updates.
% Prompt-level interventions, including mental-state reminders, perspective taking~\citep{wilf2024think}, and symbolic belief tracking~\citep{sclar-etal-2023-minding}, provide a first test of recoverability. These interventions test whether models can better apply mental-state information without changing their parameters. We then compare two ways of using answered support questions: presenting them as examples through in-context learning (ICL), or using them to update parameters through BSTTT.
To investigate parameter-based recovery, we introduce \textbf{Belief-State Test-Time Training (BSTTT)}, which adapts lightweight LoRA modules~\citep{hu2022lora} on answered support questions from each test episode before evaluating a held-out query. We compare a ToM-aligned \emph{action-reconstruction} objective with generic \emph{next-token prediction} under the same adapter configuration and update budget, and compare BSTTT with \emph{matched ICL} using the same answered supports.\\
We evaluate three open-weight instruction-tuned models (Qwen-2.5-7B, Llama-3.1-8B, and GPT-oss-20B) across four ToM benchmarks: SimpleToM~\citep{gu2026simpletom}, Hi-ToM~\citep{wu-etal-2023-hi}, OpenToM~\citep{xu-etal-2024-opentom}, and DynToM~\citep{xiao-etal-2025-towards}. SimpleToM directly separates mental-state recognition, behavior prediction, and judgment. OpenToM and DynToM extend the analysis to multihop inference and temporal belief tracking, respectively, while Hi-ToM provides a complementary test of higher-order reasoning.\\
Three findings emerge. First, \textbf{a substantial part of the explicit-to-applied gap is recoverable}, supporting H2's prediction that standard inference understates achievable performance. On SimpleToM, Llama's behavior prediction accuracy increases from 28.9\% to 63.2\% with a mental-state reminder and to 69.7\% with Reminder+BSTTT. Overall, input-level assistance and weight updates can contribute jointly to recovery.

Second, \textbf{weight updates help, but substantial recovery does not require them}. With the same answered supports, ICL can outperform BSTTT: on OpenToM, Llama reaches 56.4\% with ICL versus 43.1\% with BSTTT. Conversely, GPT-oss performs better with BSTTT, reaching 36.0\% versus 28.9\% with ICL. Thus, recovery depends both on the support answers and on how the model uses them; parameter adaptation is beneficial in some settings but is not a universal prerequisite for improved applied reasoning.

Third, \textbf{recovery is selective rather than uniform}. On SimpleToM, Llama's ICL behavior accuracy rises from 28.9\% to 66.0\%, while judgment accuracy falls from 26.2\% to 14.0\%. DynToM also shows partial recovery: Llama improves from 19.7\% to 32.6\% with ICL, leaving substantial remaining error. These results support recoverable competence in specific tasks while leaving the capacity limitations proposed by H1 unresolved. Aggregate improvements therefore cannot establish reliable applied ToM across competencies.

Our contributions are as follows:
\begin{enumerate}
    % \item \textbf{Evaluation framework:}
    % We operationalize the explicit-to-applied ToM distinction
    % (Eq.~\ref{eq:gap}) across benchmarks, exposing gaps hidden
    % by aggregate scoring.

    \item \textbf{Diagnostic protocol:}
    We introduce BSTTT and compare ToM-aligned and generic
    adaptation objectives, matched ICL, and structured prompting
    to investigate the capacity-limitation and latent-competence
    hypotheses through controlled intervention comparisons.

    \item \textbf{Cross-benchmark empirical analysis:}
    Across three model families and four benchmarks, we show
    that substantial recovery is possible without weight updates,
    that BSTTT provides additional gains in some settings, and
    that recovery varies across competencies, with persistent
    errors in temporal belief tracking.

    \item \textbf{Actionable recommendations:}
    We provide concrete guidelines for future ToM evaluation:
    separately report explicit and applied accuracy, include
    temporal reasoning as a distinct evaluation axis, and
    compare prompting, matched ICL, and weight adaptation
    to characterize recoverable competence.
\end{enumerate}

\section{The Explicit-to-Applied Theory of Mind Gap}
\label{sec:gap}

We begin by formalizing the distinction that motivates our evaluation. Consider a ToM benchmark $\mathcal{B}$ of question-answer pairs $(q_i, a_i)$ grounded in narrative social scenarios.
We partition $\mathcal{B}$ into two subsets:\\
\textbf{Explicit questions} $\mathcal{B}_\mathrm{E}$: \emph{Can the model identify an agent's mental state?} These directly probe an agent's beliefs, emotions, or intentions.\\
\textbf{Applied questions} $\mathcal{B}_\mathrm{A}$: \emph{Can the model use that mental state to predict behavior, anticipate consequences, or evaluate actions?}. These require using mental-state information together with relevant contextual, commonsense, or causal knowledge.\\
\noindent Let $\mathrm{acc}(\mathcal{M}, \mathcal{B}_\mathrm{E})$ and $\mathrm{acc}(\mathcal{M}, \mathcal{B}_\mathrm{A})$ denote a model's accuracy. The \textbf{explicit-to-applied gap} is:
\begin{equation}
    \Delta_\mathrm{gap}(\mathcal{M}, \mathcal{B}) \;=\; \mathrm{acc}(\mathcal{M}, \mathcal{B}_\mathrm{E}) \;-\; \mathrm{acc}(\mathcal{M}, \mathcal{B}_\mathrm{A})
    \label{eq:gap}
\end{equation}
A positive $\Delta_\mathrm{gap}$ indicates higher accuracy on explicit than applied questions. A gap near zero indicates similar accuracy across the two subsets, which may reflect either strong or weak performance; we therefore report both accuracies alongside the gap.
\subsection{Operationalization Across Benchmarks}
\label{sec:gap_operationalization}
\begin{table*}[t]
    \caption{Each benchmark decomposes into explicit (mental-state recognition) and applied (behavioral inference) subsets. Hi-ToM provides a complementary test of higher-order reasoning recovery without a paired explicit subset.
    ($|\mathcal{B}|$: Total number of questions; $C$ = number of answer choices;
    $|\mathcal{S}|$: support set size per episode.)}
    \label{tab:gap_operationalization}
    \small
    \centering
    \begin{tabular}{@{}lllllll@{}}
        \toprule
        \textbf{Benchmark} & \textbf{Explicit ($\mathcal{B}_\mathrm{E}$)} & 
        \textbf{Applied ($\mathcal{B}_\mathrm{A}$)} & $C$ & $|\mathcal{B}|$ & 
        \textbf{Regime} & $|\mathcal{S}|$ \\
        \midrule
        SimpleToM & Mental-state & Behavior, Judgment & 2 & 3,441 & 
        Static, 1\textsuperscript{st}-order & 1 \\
        DynToM & Understand (A) & Transform (C,D) & 7--8 & 3,000 & 
        Temporal, multi-step & $\leq$4 \\
        Hi-ToM & \emph{(indirect)} & Recursive belief, location & 15 & 1,200 & 
        Static, higher-order & $\leq$4 \\
        OpenToM & Attitude & Multihop first-order & 3--6 & 2,000 & 
        Static, multi-agent & $\leq$4 \\
        \bottomrule
    \end{tabular}
\end{table*}

Table~\ref{tab:gap_operationalization} summarizes our
operationalization across four ToM regimes; examples appear in Appendix~\ref{app:examples}. \textbf{SimpleToM}~\citep{gu2026simpletom} provides parallel mental-state, behavior, and judgment questions for each scenario, making the gap directly measurable. \textbf{DynToM}~\citep{xiao-etal-2025-towards} contrasts state identification with reasoning about within-scenario state relations and cross-scenario transformations. \textbf{Hi-ToM}~\citep{wu-etal-2023-hi} evaluates object locations and recursively nested beliefs. Without a paired recognition application decomposition \textbf{OpenToM}~\citep{xu-etal-2024-opentom} contrasts attitude inference, used as an explicit proxy, with multihop first-order belief reasoning.

% Table~\ref{tab:gap_operationalization} maps each benchmark to its explicit and applied subsets. Concrete examples of each question format are provided in Appendix \ref{app:examples}.\\
% The four benchmarks span qualitatively different ToM regimes.
% \textbf{SimpleToM} \citep{gu2026simpletom} provides the cleanest operationalization: each scenario generates parallel questions about the same social event, one for mental-state identification, one for behavior prediction, one for judgment, making the gap directly measurable.
% \textbf{DynToM} \citep{xiao-etal-2025-towards} extends the distinction temporally: understanding questions probe an agent's current mental state; transformation questions require tracking how that state evolves across a five-event trajectory and manifests in behavior.
% \textbf{Hi-ToM} \citep{wu-etal-2023-hi} evaluates object
% locations and recursively nested beliefs, such as where
% Alice thinks Bob believes an object is located. It provides
% a complementary test of higher-order reasoning recovery.
% Because the evaluated question types do not provide a
% paired recognition application decomposition, we report
% accuracy and intervention gains without assigning an
% explicit-to-applied gap.
% \textbf{OpenToM} \citep{xu-etal-2024-opentom} separates attitude inference (closer to explicit recognition) from multihop first-order reasoning that chains mental-state attributions across agents to reach behavioral conclusions.

\subsection{Diagnostic Question}
\label{sec:diagnostic_question}
A positive $\Delta_\mathrm{gap}$ under standard inference motivates two hypotheses:\\
\textbf{H1 (Capacity limitation):} The model lacks some reasoning capabilities needed to translate mental-state recognition into reliable application.\\
\textbf{H2 (Latent competence):} The model possesses relevant social knowledge but fails to consistently apply it under standard inference.\\
We investigate these hypotheses through prompting, matched ICL, and BSTTT, comparing recovery from explicit belief cues, answered supports, and parameter updates. Recovery shows that standard inference understates achievable performance, but does not establish that all required knowledge was already encoded in the model. Persistent errors identify limitations of the tested interventions without proving an intrinsic capacity deficit.

% \subsection{Diagnostic Question}
% \label{sec:diagnostic_question}
% A positive $\Delta_\mathrm{gap}$ under frozen inference admits two competing explanations: \\
% \textbf{H1 (Capacity deficit)}: The model lacks internal representations needed to bridge from recognition to prediction. The gap is irreducible without additional training or architectural change. \\
% \textbf{H2 (Latent competence)}: The model's parameters encode sufficient social knowledge, but standard decoding fails to activate the belief-to-behavior mapping. The gap is inference-time failure. \\
% Prompt-level interventions cannot resolve this distinction: they modify what the model \emph{sees} but do not reveal whether the model's \emph{parameters} encode the requisite knowledge. Resolving H1 versus H2 requires a controlled intervention on model parameters.

\section{Belief-State Test-Time Training}
\label{sec:bsttt}
We introduce Belief-State Test-Time Training (BSTTT), it updates LoRA parameters on an episode's support set and uses the adapted model to score candidate answers for a held-out query. The backbone parameters remain fixed, and the LoRA parameters are reset before each episode.
\subsection{Episode Construction}
\label{sec:episodic}
BSTTT operates episodically. Each \textbf{episode} $e = (\mathcal{S}, q)$ consists of a 
\textbf{support set} $\mathcal{S} = \{(x_j, a_j^*)\}_{j=1}^{|\mathcal{S}|}$ of answered 
questions and a held-out \textbf{query} $q = (x_q, a_q^*)$, with supports drawn from the query's scenario family in SimpleToM, narrative group in Hi-ToM and OpenToM, or trial in DynToM. No data external to the benchmark is used. The held-out query is excluded from its support set, and its reference label is used only for evaluation. Support set sizes are benchmark-specific and reported as $|\mathcal{S}|$ in Table~\ref{tab:gap_operationalization}; Full episode construction details are in Appendix~\ref{app:benchmark_descriptions}. 

\subsection{Adaptation Objectives}
\label{sec:objectives}
We compare two objectives using the same model, support items, LoRA configuration, optimizer, and number of gradient steps.

\noindent\textbf{Action Reconstruction (AR):}
AR minimizes cross-entropy over candidate answer scores.
For support item $j$, let $s_{j,c} = \log p(o_{j,c} \mid x_j;\theta+\Delta\theta)$ denote the log-likelihood of candidate $c$, and let $y_j$ index the correct answer $a_j^*$ among $C_j$ candidates:
\begin{equation}
    \mathcal{L}_\mathrm{AR}
    = -\frac{1}{|\mathcal{S}|}
    \sum_{j=1}^{|\mathcal{S}|}
    \log
    \frac{\exp(s_{j,y_j})}
         {\sum_{c=1}^{C_j}\exp(s_{j,c})}.
    \label{eq:ar_loss}
\end{equation}

\noindent\textbf{Next-Token Loss (NTL):} NTL applies causal language modeling to the narratives of the same support items. Let $z_j=(z_{j,1},\ldots,z_{j,T_j})$ denote the tokenized support narrative and $T=\sum_j(T_j-1)$ the total number of prediction targets, excluding padding:
\begin{equation}
    \mathcal{L}_\mathrm{NTL}
    = -\frac{1}{T}
    \sum_{j=1}^{|\mathcal{S}|}
    \sum_{t=2}^{T_j}
    \log p(z_{j,t}\mid z_{j,<t};\theta+\Delta\theta).
    \label{eq:ntl_loss}
\end{equation}
This comparison measures the benefit of answer-supervised
adaptation over generic modeling of the support narratives.

% \subsection{Adaptation Objectives}
% \label{sec:objectives}

% The diagnostic power of BSTTT lies in comparing two adaptation objectives under identical condition, same model, same LoRA parameters, same support data, same gradient steps. \\
% \textbf{Action Reconstruction (AR):}
% The ToM-aligned condition trains the model to predict the behaviorally correct answer. Given a support item $(x_j, a_j^*)$ with $C$ multiple-choice options $\{o_1, \ldots, o_C\}$, let $s_c = \log p(o_c \mid x_j;\, \theta + \Delta\theta)$ denote the log-likelihood of option $o_c$ under the adapted model:
% \begin{equation}
%     \mathcal{L}_\mathrm{AR} = -\frac{1}{|\mathcal{S}|} \sum_{j=1}^{|\mathcal{S}|} \log \frac{\exp(s_{a_j^*})}{\sum_{c=1}^{C} \exp(s_c)}
%     \label{eq:ar_loss}
% \end{equation} \\
% \textbf{Next-Token Loss (NTL):}
% The control condition applies standard causal language modeling:
% \begin{equation}
%     \mathcal{L}_\mathrm{NTL} = -\frac{1}{T} \sum_{t=1}^{T} \log p(x_t \mid x_{<t};\, \theta + \Delta\theta)
%     \label{eq:ntl_loss}
% \end{equation}
% If only AR closes the gap, the recovery is specific to the belief-to-behavior mapping, supporting~\textbf{H2}.

\subsection{Implementation}
\label{sec:adaptation}
We attach rank-$r$ LoRA adapters~\citep{hu2022lora} to
the top 25\% of transformer layers, keeping the backbone
parameters fixed. The adapters are initialized with a
zero effective weight update. Before each episode,
adapter parameters are restored to their initial values
and optimizer state is cleared. Adaptation uses $K$
gradient steps with AdamW~\citep{loshchilov2018decoupled}.
Default hyperparameters are $r{=}8$, $\alpha{=}16$,
$K{=}3$, and $\eta{=}10^{-4}$, with zero weight decay. Full configuration, run-specific settings, and scoring
details appear in Appendix~\ref{app:bsttt_details}.

\noindent\textbf{Composition with prompting:}
The adapted model scores candidate answers conditioned
on the prompt-transformed query:
\begin{equation}
\begin{aligned}
    \hat{a} = \argmax_{o_c}\;&
    \log p\!\bigl(o_c \mid f_\mathrm{prompt}(x_q);\\
    &\theta + \Delta\theta_\mathrm{BSTTT}\bigr).
\end{aligned}
\label{eq:composition}
\end{equation}
Here, $f_\mathrm{prompt}$ denotes the selected prompting
procedure, including MS-Reminder, SimToM, SymbolicToM,
or Scratchpad. Method-specific support transformations
and intermediate-state generation are detailed in
Appendix~\ref{app:bsttt_details}.
% At the start of each episode, we attach rank-$r$ LoRA adapters~\citep{hu2022lora} to the top 25\% of transformer layers and initialize them to zero.
% Adaptation uses $K$ gradient steps with AdamW~\citep{loshchilov2018decoupled}; both LoRA weights and optimizer state are reset per episode.
% We keep this configuration fixed across AR and NTL comparisons so that any recovery can be attributed to the supervision signal rather than benchmark-specific tuning.
% Default hyperparameters: $r{=}8$, $\alpha{=}16$, $K{=}3$, $\eta{=}10^{-4}$ (robustness to adaptation budget is examined in Appendix~\ref{app:ablations}). Full LoRA configuration, episodic reset protocol, and scoring details are in Appendix\ref{app:bsttt_details}. \\
% \textbf{Composition with prompting:}
% Because BSTTT modifies weights and prompting modifies inputs, they compose naturally:
% \begin{equation}
%     \hat{a} = \argmax_{c} \;\; \log p\!\left(o_c \;\middle|\; f_\mathrm{prompt}(x_q);\; \theta + \Delta\theta_\mathrm{BSTTT}\right)
%     \label{eq:composition}
% \end{equation}
% where $f_\mathrm{prompt}$ denotes any input transformation (CoT, SimToM, SymbolicToM, or Scratchpad).

\section{Experimental Setup}
\label{sec:setup}
\textbf{Benchmarks \& Model Details:} We evaluate across four benchmarks (Table~\ref{tab:gap_operationalization}) covering static/temporal scenarios, first/higher-order belief, and recognition/application. Full benchmark descriptions including question taxonomies and our operationalization choices are in Appendix \ref{app:benchmark_descriptions}. \\
We evaluate three instruction-tuned open-weight models: \textbf{Qwen-2.5-7B-Instruct}~\citep{qwen2.5}, \textbf{Llama-3.1-8B-Instruct}~\citep{grattafiori2024llama}, and \textbf{GPT-oss-20B}~\citep{openai2025gptoss120bgptoss20bmodel}\\
\textbf{Method Configurations:}
\begin{table*}[t]
    \caption{Method configurations and their diagnostic purpose. See Table~\ref{tab:appendix_method_map} for benchmark-specific instantiations}
    \label{tab:methods}
    \centering
    \small
    \begin{tabular}{@{}lcccl@{}}
        \toprule
        \textbf{Method} & \textbf{Adapts $\theta$?} & \textbf{Prompt} & \textbf{Objective} & \textbf{What it tests} \\
        \midrule
        \multicolumn{5}{@{}l}{\textit{Frozen baselines}} \\
        Frozen & \ding{55} & Vanilla & \text{---} & Baseline gap \\
        CoT & \ding{55} & Step-by-step & \text{---} & Reasoning elicitation \\
        \multicolumn{5}{@{}l}{\textit{Structured prompting}} \\
        Scratchpad & \ding{55} & Belief scaffold & \text{---} & Externalizing tracking \\
        SimToM & \ding{55} & 2-stage perspective & \text{---} & Perspective filtering \\
        SymbolicToM & \ding{55} & Symbolic tracker & \text{---} & Explicit state representation\\
        \multicolumn{5}{@{}l}{\textit{In-context learning}} \\ Matched ICL & \ding{55} & Answered supports & --- & Support use without weight updates \\
        Cross-scenario ICL & \ding{55}
& Cross-scenario demonstrations & ---
& Support use without weight updates \\
        \multicolumn{5}{@{}l}{\textit{Test-time training}} \\
        BSTTT-NTL & \ding{51} & Vanilla & NTL (Eq.~\ref{eq:ntl_loss}) & Generic exposure \\
        BSTTT-AR & \ding{51} & Vanilla & AR (Eq.~\ref{eq:ar_loss}) & ToM-aligned \\
        \multicolumn{5}{@{}l}{\textit{Composed (BSTTT + prompt)}} \\
        Scratchpad{+}BSTTT & \ding{51} & Scratchpad & AR & Weight + prompting \\
        SimToM{+}BSTTT & \ding{51} & SimToM & AR & Weight + prompting \\
        SymbolicToM{+}BSTTT & \ding{51} & SymbolicToM & AR & Weight + prompting \\
        \bottomrule
    \end{tabular}
\end{table*}
We compare 12 configurations (Table~\ref{tab:methods}).
Matched ICL presents the same answered supports used by
BSTTT as demonstrations, keeping model parameters fixed. We also evaluate cross-scenario ICL using answered supports from other scenario families or trials (Appendix~\ref{app:cross_scenario_icl}). BSTTT settings follow \S\ref{sec:adaptation}; prompt templates and benchmark-specific instantiations appear in Appendix~\ref{app:method_details} and
Table~\ref{tab:appendix_method_map}. All experiments use three seeds $\{42,43,44\}$. We report mean accuracy, with 95\% bootstrap confidence intervals based on 2{,}000 resamples in Appendix~\ref{app:bootstrap}.

\section{Results} \label{sec:results} 

\begin{figure*}[t]
\small
\centering
\includegraphics[width=\linewidth]{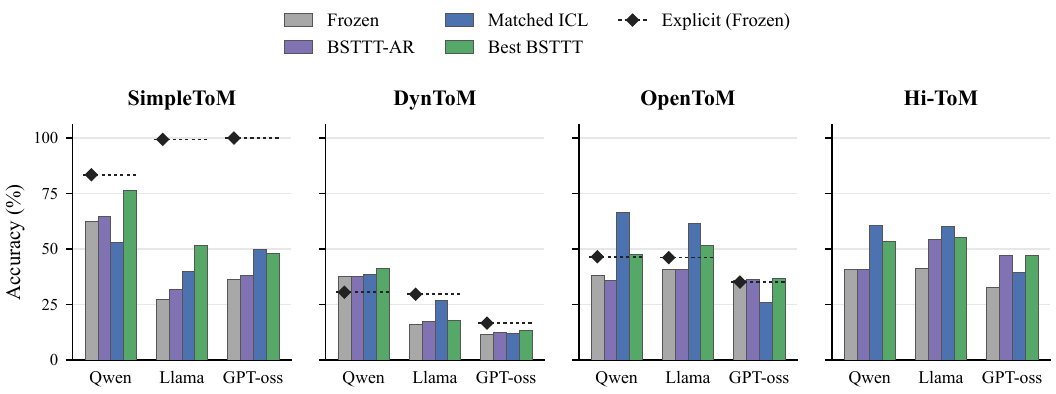}
\caption{\textbf{Explicit-to-applied gap and recovery across benchmarks} Bars show accuracy pooled over applied questions; diamonds indicate frozen explicit accuracy. Their difference from the frozen bars gives the baseline gap. Hi-ToM shows overall accuracy without an explicit reference.``Best BSTTT'' shows the applied accuracy of the configuration selected by overall accuracy in Table~\ref{tab:summary}. Intervention gains are measured relative to the frozen bars; evaluation settings follow Table~\ref{tab:summary}.}
\label{fig:gap_recovery}
\end{figure*}

We examine the gap and its recoverability
(\S\ref{sec:domain_specific}), compare interventions
(\S\ref{sec:composability}), and analyze variation across reasoning tasks (\S\ref{sec:structural_variation}) and failure modes (\S\ref{sec:error_analysis}). Table~\ref{tab:summary} provides a cross-model overview; Figure~\ref{fig:gap_recovery} shows the gap and intervention gains, with numerical values in Table~\ref{tab:fig2_gap_recovery_values}. Full results appear in Appendix~\ref{app:consolidated_results}.

\begin{table*}[t]
    \caption{\textbf{Cross-model summary.}
    Accuracy (\%). ``Best prompt'' selects the strongest
    prompting method without demonstrations or weight updates;
    ``Best BSTTT'' selects the strongest BSTTT configuration,
    including prompting compositions.
    Bold indicates the highest listed accuracy per model and
    benchmark. $\Delta_{\mathrm{best}}$ is the highest listed
    accuracy minus Frozen, in percentage points.
    All accuracies are averaged over three seeds.
    Run-specific settings are reported in
    Appendix~\ref{app:bsttt_details}.}
    \label{tab:summary}
    \small
    \centering
    \begin{tabular}{@{}llcccc@{}}
        \toprule
        \textbf{Model} & \textbf{Condition} &
        \textbf{SimpleToM} & \textbf{Hi-ToM} &
        \textbf{OpenToM} & \textbf{DynToM} \\
        \midrule
        \multirow{6}{*}{Llama-3.1-8B}
            & Frozen       & 51.5 & 41.4 & 42.5 & 19.7 \\
            & Best prompt  & 65.5\textsuperscript{h}
                           & 47.5\textsuperscript{b}
                           & 48.8\textsuperscript{b}
                           & 24.5\textsuperscript{g} \\
            & BSTTT-AR     & 54.2 & 54.4 & 43.1 & 20.7 \\
            & Matched ICL  & 59.8 & \textbf{60.3}
                           & \textbf{56.4} & \textbf{32.6} \\
            & Best BSTTT   & \textbf{67.7}\textsuperscript{i}
                           & 55.3\textsuperscript{d}
                           & 50.0\textsuperscript{d}
                           & 22.2\textsuperscript{c} \\
            & $\Delta_{\mathrm{best}}$
                           & +16.2 & +18.9 & +13.9 & +12.9 \\
        \midrule
        \multirow{6}{*}{Qwen-2.5-7B}
            & Frozen       & 69.4 & 40.7 & 40.5 & 35.7 \\
            & Best prompt  & 77.0\textsuperscript{h}
                           & 43.8\textsuperscript{a}
                           & 45.6\textsuperscript{e}
                           & 38.4\textsuperscript{a} \\
            & BSTTT-AR     & 70.8 & 40.7 & 40.5 & 35.4 \\
            & Matched ICL  & 62.4 & \textbf{60.6}
                           & \textbf{60.4} & \textbf{39.9} \\
            & Best BSTTT   & \textbf{82.5}\textsuperscript{i}
                           & 53.7\textsuperscript{c}
                           & 46.7\textsuperscript{f}
                           & 38.8\textsuperscript{c} \\
            & $\Delta_{\mathrm{best}}$
                           & +13.1 & +19.9 & +19.9 & +4.2 \\
        \midrule
        \multirow{6}{*}{GPT-oss-20B}
            & Frozen       & 57.6 & 32.6 & 35.6 & 13.0 \\
            & Best prompt  & 64.1\textsuperscript{h}
                           & 34.8\textsuperscript{a}
                           & 35.8\textsuperscript{a}
                           & \textbf{14.4}\textsuperscript{g} \\
            & BSTTT-AR     & 58.9 & \textbf{47.3} & 36.0 & 13.2 \\
            & Matched ICL  & \textbf{66.7} & 39.5 & 28.9 & 13.3 \\
            & Best BSTTT   & 65.4\textsuperscript{i}
                           & \textbf{47.3}
                           & \textbf{36.3}\textsuperscript{c}
                           & 14.1\textsuperscript{c} \\
            & $\Delta_{\mathrm{best}}$
                           & +9.1 & +14.7 & +0.7 & +1.4 \\
        \bottomrule
    \end{tabular}

    \par\smallskip
    {\footnotesize
    \textsuperscript{a}Scratchpad;
    \textsuperscript{b}SymbolicToM;
    \textsuperscript{c}Scratchpad+BSTTT
    (Scratchpad-TTT on DynToM);
    \textsuperscript{d}SymbolicToM+BSTTT;
    \textsuperscript{e}SimToM;
    \textsuperscript{f}SimToM+BSTTT;
    \textsuperscript{g}CoT;
    \textsuperscript{h}MS-Reminder;
    \textsuperscript{i}MS-Reminder+BSTTT.
    }
\end{table*}

\subsection{Finding 1: The Gap Is Substantial and Partially Recoverable}
\label{sec:domain_specific}

\begin{table*}[t]

    \caption{\textbf{AR vs.\ NTL: the diagnostic contrast.} Both conditions use identical LoRA configuration, support data, and gradient budget. $\Delta$(AR$-$NTL) reports the accuracy
    difference between answer-supervised adaptation and
    narrative modeling. All accuracies are averaged over three seeds.}
    \label{tab:ar_vs_ntl}
    \centering
    \small
    \begin{tabular}{@{}llcccr@{}}
        \toprule
        \textbf{Model} & \textbf{Benchmark} &
        \textbf{Frozen} & \textbf{BSTTT-NTL} &
        \textbf{BSTTT-AR} &
        $\boldsymbol{\Delta}$\textbf{(AR$-$NTL)} \\
        \midrule
        \multirow{4}{*}{Llama-8B}
            & SimpleToM & 51.5 & 51.7 & 54.2 & +2.5 \\
            & Hi-ToM    & 41.4 & 41.4 & 54.4 & +13.0 \\
            & OpenToM   & 42.5 & 42.7 & 43.1 & +0.4 \\
            & DynToM    & 19.7 & 19.9 & 20.7 & +0.8 \\
        \midrule
        \multirow{4}{*}{Qwen-7B}
            & SimpleToM & 69.4 & 69.0 & 70.8 & +1.8 \\
            & Hi-ToM    & 40.7 & 40.7 & 40.7 & 0.0 \\
            & OpenToM   & 40.5 & 40.5 & 40.5 & 0.0 \\
            & DynToM    & 35.7 & 35.7 & 35.4 & -0.3 \\
        \midrule
        \multirow{4}{*}{GPT-oss-20B}
            & SimpleToM & 57.6 & 56.9 & 58.9 & +2.0 \\
            & Hi-ToM    & 32.6 & 31.6 & 47.3 & +15.7 \\
            & OpenToM   & 35.6 & 35.1 & 36.0 & +0.9 \\
            & DynToM    & 13.0 & 13.2 & 13.2 & 0.0 \\
        \bottomrule
    \end{tabular}
\end{table*}

\noindent\textbf{The gap landscape:}
Figure~\ref{fig:gap_recovery} and
Table~\ref{tab:fig2_gap_recovery_values} show substantial variation across benchmarks. On SimpleToM, frozen explicit-to-applied gaps range from 21.0 to 71.9~pp, with applied accuracy pooling behavior and judgment questions. Llama exhibits the largest difference: 99.4\% mental-state recognition versus 27.5\% applied accuracy. OpenToM gaps are smaller ($-0.8$ to 8.5~pp), while DynToM includes a reversal for Qwen, whose transformation accuracy exceeds understanding accuracy by 7.3~pp. Thus, the recognition--application disparity is pronounced on SimpleToM but varies with the model and reasoning task. Hi-ToM provides a separate measure of higher-order reasoning recovery without an explicit reference subset.\\
\noindent\textbf{Recovery extends beyond weight adaptation:}
On SimpleToM, Llama's applied accuracy increases from
27.5\% to 51.7\% with MS-Reminder+BSTTT and to 40.0\%
with matched ICL. Contextual examples also improve
OpenToM multihop accuracy from 38.0\% to 66.5\% for
Qwen and from 40.9\% to 61.5\% for Llama. On DynToM, Llama's transformation accuracy rises from
15.9\% to 26.8\% with ICL, revising the earlier picture
of minimal temporal recovery. These results support
H2's prediction that some failures under standard
inference are recoverable, while showing that recovery
can occur through both contextual assistance and
parameter adaptation.\\
\noindent\textbf{Is the gap a capacity limitation or recoverable competence?}
Table~\ref{tab:ar_vs_ntl} compares answer-supervised adaptation with narrative modeling under the same adapter configuration and update budget. On Hi-ToM, AR exceeds NTL by 13.0~pp for Llama and 15.7~pp for GPT-oss, whereas NTL provides no improvement over frozen inference. On SimpleToM, AR's overall advantage is smaller (1.8--2.5~pp).
These results show that the adaptation objective and supplied supervision affect recovery. However, the comparison alone does not establish that the required knowledge was already encoded in the original parameters. On DynToM, AR--NTL differences range from $-0.3$ to $+0.8$~pp; the ICL gains reported above show that limited recovery under these adaptation objectives does not establish H1.\\
\noindent\textbf{Separating recovery from the contribution of weight updates:}
Figure~\ref{fig:gap} and Table~\ref{tab:simpletom_subtypes}
decompose SimpleToM performance by question type.
Llama's behavior accuracy rises from 28.9\% to 69.7\%
with MS-Reminder+BSTTT, while recognition remains near
ceiling. This 40.8-point improvement supports recoverable
application under H2, but its attribution requires the
component comparisons: MS-Reminder alone reaches 63.2\%,
whereas BSTTT-AR alone reaches 37.1\%.
Recovery also depends on the applied task. GPT-oss
judgment accuracy rises from 3.1\% to 18.7\% with
MS-Reminder+BSTTT and to 26.8\% with matched ICL.
Thus, the strongest intervention can differ across
competencies even within the same model and benchmark.\\
\noindent\textbf{Where AR alone provides little benefit:}
Qwen's accuracy remains unchanged under BSTTT-AR:
40.7\% on Hi-ToM and 40.5\% on OpenToM
(Table~\ref{tab:summary}). Nevertheless, prompting
compositions reach 53.7\% and 46.7\%, respectively,
and matched ICL reaches 60.6\% and 60.4\%. Failure to improve with AR alone therefore does not imply that the task is beyond the model's achievable performance. These outcomes motivate the direct comparison of contextual assistance, adaptation, and their composition in \S\ref{sec:composability}.

\subsection{Finding 2: Recovery Depends on the Intervention}
\label{sec:composability}
\begin{table*}[t]
\caption{\textbf{Component comparison for prompting compositions.} Accuracy (\%) averaged over three seeds.
$P$: the selected prompting method; $B$: BSTTT-AR.
$\Delta(B{+}P-P)$: additional gain from adaptation over prompting alone. $\Delta(B{+}P-B)$: additional gain from prompting over adaptation alone. Differences are reported in percentage points.}
\label{tab:composability}
\centering
\small
\begin{tabular}{@{}llcccrr@{}}
\toprule
\textbf{Model} & \textbf{Benchmark} &
$\boldsymbol{P}$ & $\boldsymbol{B}$ &
$\boldsymbol{B{+}P}$ &
$\boldsymbol{\Delta(B{+}P-P)}$ &
$\boldsymbol{\Delta(B{+}P-B)}$ \\
\midrule
\multirow{4}{*}{Llama-8B}
& SimpleToM\textsuperscript{a}
    & 65.5 & 54.2 & 67.7 & +2.2 & +13.5 \\
& Hi-ToM\textsuperscript{b}
    & 47.5 & 54.4 & 55.3 & +7.8 & +0.9 \\
& OpenToM\textsuperscript{b}
    & 48.8 & 43.1 & 50.0 & +1.2 & +6.9 \\
& DynToM\textsuperscript{c}
    & 20.8 & 20.7 & 22.2 & +1.4 & +1.5 \\
\midrule
\multirow{4}{*}{Qwen-7B}
& SimpleToM\textsuperscript{a}
    & 77.0 & 70.8 & 82.5 & +5.5 & +11.7 \\
& Hi-ToM\textsuperscript{c}
    & 43.8 & 40.7 & 53.7 & +9.9 & +13.0 \\
& OpenToM\textsuperscript{d}
    & 45.6 & 40.5 & 46.7 & +1.1 & +6.2 \\
& DynToM\textsuperscript{c}
    & 38.4 & 35.4 & 38.8 & +0.4 & +3.4 \\
\midrule
\multirow{4}{*}{GPT-oss-20B}
& SimpleToM\textsuperscript{a}
    & 64.1 & 58.9 & 65.4 & +1.3 & +6.5 \\
& Hi-ToM\textsuperscript{c}
    & 34.8 & 47.3 & 46.5 & +11.7 & -0.8 \\
& OpenToM\textsuperscript{c}
    & 35.8 & 36.0 & 36.3 & +0.5 & +0.3 \\
& DynToM\textsuperscript{c}
    & 14.3 & 13.2 & 14.1 & -0.2 & +0.9 \\
\bottomrule
\end{tabular}

\par\smallskip
{\footnotesize
Selected prompts:
\textsuperscript{a}MS-Reminder;
\textsuperscript{b}SymbolicToM;
\textsuperscript{c}Scratchpad
(Scratchpad-Frozen and Scratchpad-TTT on DynToM);
\textsuperscript{d}SimToM.
}
\end{table*}

Three patterns emerge.

\noindent\textbf{Answered examples can outperform weight updates:}
Table~\ref{tab:summary} shows that contextual use of supports can yield substantial gains without adaptation. On OpenToM, matched ICL exceeds BSTTT-AR by 13.3~pp for Llama and 19.9~pp for Qwen, whereas GPT-oss favors BSTTT-AR by 7.1~pp. Hi-ToM exhibits the same model-dependent ordering: ICL reaches 60.3\% and 60.6\% for Llama and Qwen, compared with 54.4\% and 40.7\% under BSTTT-AR; GPT-oss instead reaches 47.3\% with BSTTT-AR versus 39.5\% with ICL. Thus, answered supports do not require weight updates to improve performance, and their more effective mode of use varies across models.\\
\noindent\textbf{Recovery with cross-scenario supports:}
On SimpleToM, GPT-OSS achieves 73.4\% behavior accuracy
and 42.2\% judgment accuracy with cross-scenario ICL
(Appendix~\ref{app:cross_scenario_icl}). These results
show that useful applied performance is possible
without demonstrations from the query's scenario
family, although they do not isolate answer-prior effects.\\
\noindent\textbf{Weight adaptation can add to prompting gains:}
Table~\ref{tab:composability} separates the contribution of
each component. On SimpleToM, composing MS-Reminder with BSTTT adds 2.2~pp for Llama, 5.5~pp for Qwen, and 1.3~pp for GPT-oss over the reminder alone. Complementarity is more pronounced for Qwen on Hi-ToM: Scratchpad achieves 43.8\%, BSTTT-AR achieves 40.7\%, and their composition reaches 53.7\%, exceeding both components. On OpenToM, Llama's SymbolicToM+BSTTT reaches 50.0\%, but the gain over SymbolicToM alone is only 1.2~pp. Improvements over BSTTT alone therefore need to be read alongside improvements over prompting alone.\\
\noindent\textbf{Composition is not uniformly beneficial:}
For GPT-oss on Hi-ToM, adding Scratchpad reduces accuracy from 47.3\% under BSTTT-AR to 46.5\%. On DynToM, Scratchpad-TTT reaches 14.1\%, below the 14.3\% achieved by Scratchpad-Frozen. These cases show that a composition can exceed one component while underperforming the other. The results support model and task-dependent benefits
from combining interventions, rather than a general
advantage of composition.

\subsection{Finding 3: Recovery Varies Across Reasoning Tasks} \label{sec:structural_variation}
The preceding findings show that recovery depends on the operation being evaluated: gains in behavior prediction can accompany regressions in judgment, and the relative benefits of ICL and adaptation vary across reasoning tasks (\S\ref{sec:domain_specific}--\ref{sec:composability}).
\noindent\textbf{DynToM: partial recovery with persistent errors.} DynToM extends this analysis to within-scenario mental-state relations and cross-scenario transformations. Matched ICL improves Llama's pooled applied accuracy from 15.9\% to 26.8\% (Figure~\ref{fig:gap_recovery}), demonstrating partial recovery while leaving substantial remaining error. 
The Scratchpad-Oracle control further examines whether providing annotated mental-state tables resolves these difficulties. Overall accuracy reaches only 42.2\% for Qwen, 22.0\% for Llama, and 14.5\% for GPT-oss. Making state information available is therefore insufficient for reliable performance under this prompting configuration. These results motivate evaluating state identification separately from reasoning about relations and changes between states and the capacity-limitation hypothesis remains unresolved. Full results appear in Table~\ref{tab:main_dyntom_datasetwise}.

\subsection{Error Analysis}
\label{sec:error_analysis}
Per-instance comparisons show that interventions succeed
on partly different questions: ICL's aggregate advantage
on OpenToM coexists with cases where BSTTT-AR is more
effective. These opposing outcomes motivate examining
corrections and regressions alongside overall accuracy.

Qualitative review suggests three recurring failure patterns:
(i)~\emph{limited support transfer}, where performance on
support questions does not transfer to the query;
(ii)~\emph{multistep inference errors}, involving incorrect
intermediate beliefs or temporal transitions; and
(iii)~\emph{perspective errors}, where predictions follow
world knowledge rather than the agent's available information.
These patterns characterize remaining failures without
establishing an intrinsic capacity limit.

\section{Recommendations for Theory of Mind Evaluation} \label{sec:recommendations}
Our findings motivate four recommendations. \textbf{R1: Decompose aggregate scores into explicit and applied subsets.} Report recognition separately from application, and distinguish applied subtypes: gains in behavior prediction can coexist with regressions in judgment. \textbf{R2: Evaluate temporal reasoning separately.} Distinguish temporal belief tracking from static inference to identify difficulties in reasoning about state changes. \textbf{R3: Compare recoverability under matched support.} Evaluate ICL and weight adaptation using the same answered supports to distinguish contextual assistance from parameter updates. These comparisons characterize recovery under the tested interventions. \textbf{R4: Evaluate compositions against both components.} Compare prompting plus adaptation with prompting alone and adaptation alone, since their additional benefit depends on the model and task.

\section{Related Work} \label{sec:related}
\textbf{ToM evaluation in LLMs:}
False-belief evaluations report strong LLM
performance~\citep{kosinski2024evaluating}, while other
studies identify failures under minor
perturbations~\citep{ullman2023large}, adversarial
probes~\citep{shapira2024clever}, and broader social
reasoning tasks~\citep{sap2022neural}. Additional background appears in Appendix~\ref{background}.
\noindent\textbf{Prompting and in-context learning for social reasoning:}
Reasoning instructions and in-context examples can improve
ToM performance~\citep{moghaddam2023boosting}.
SimToM~\citep{wilf2024think} decomposes perspective-taking;
SymbolicToM~\citep{sclar-etal-2023-minding} tracks beliefs
symbolically; SimpleToM demonstrates gains from mental-state
reminders~\citep{gu2026simpletom}.
We compare prompting, ICL with matched answered supports,
and BSTTT to examine recovery through contextual
information, parameter updates, and their composition.
\noindent\textbf{Test-time training and episodic adaptation:}
Test-time training adapts parameters during evaluation,
originally through self-supervision on test
inputs~\citep{Sun2019TestTimeTW, NEURIPS2021_b618c321}.
For language models, In-Place TTT updates existing MLP
weights using a language-modeling-aligned
objective~\citep{feng2026inplace}, while TTT-Discover
uses test-time RL for problem solving~\citep{2026arXiv260116175Y}. BSTTT performs adaptation with AR--NTL and ICL comparisons examins the adaptation objective and how support information is used.

\section{Conclusion}
\label{sec:conclusion}
Aggregate ToM scores obscure differences between recognizing mental states and applying them. Across four benchmarks, we examine recovery through prompting, ICL, and our proposed BSTTT method. Answered supports enable substantial recovery without weight updates, while adaptation provides additional benefits in specific model-task settings. Cross-scenario ICL also supports applied performance, showing that useful demonstrations need not come from the query's scenario family. Recovery is selective: gains in behavior prediction can accompany regressions in judgment. On DynToM, partial recovery and persistent errors under annotated-state prompting leave capacity limitations unresolved. These findings motivate separate reporting of recognition and applied subtypes, together with matched-support and component comparisons to establish which failures are recoverable under what conditions.

\section*{Limitations}
\label{sec:limitations}
Three limitations bound the scope of our conclusions. First, all experiments use instruction-tuned models at the 7–20B scale. Whether the explicit-to-applied gap persists, narrows, or takes different forms at 70B+ or in frontier models remains an open question, though BSTTT is agnostic to scale and applies to any model supporting gradient-based adaptation. Second, all four benchmarks use multiple-choice formats; open-ended generation may reveal different gap, particularly for applied reasoning that requires constructing rather than selecting a behavioral prediction. Third, BSTTT's episodic reset, which is essential for isolating per-scenario adaptation signal is structurally mismatched to temporally extended tasks like DynToM, where belief states evolve across episodes. Continual adaptation without reset is a natural direction for this regime. Our finding that applied ToM is often latent than absent means that standard frozen evaluation systematically underestimates what models can do but also what they reliably do in deployment may be considerably less. A model that reasons correctly about mental states under targeted adaptation presents a deceptive capability profile in social contexts. Decomposed evaluation (section: \ref{sec:recommendations}), is a concrete step towards making this gap visible.

\section*{Ethical Considerations}
Our evaluation uses existing Theory of Mind benchmarks.
Performance on these tasks should not be interpreted
as evidence of human-like social understanding or
readiness for deployment in therapy, education, or
personal assistance. Improvements under test-time
interventions remain dependent on the model, task,
and available support information. Applications
involving judgments about people's mental states
require evaluation beyond these benchmarks.

\section*{AI Assistants Usage}
AI assistants are used in a limited and clearly defined manner in this work. AI-based writing assistants are used during manuscript preparation for grammar checking, language polishing, and improving clarity. These tools do not contribute to the scientific content, experimental design, analysis, or conclusions of the paper.

\bibliography{custom}

@article{Premack_Woodruff_1978, title={Does the chimpanzee have a theory of mind?}, volume={1}, DOI={10.1017/S0140525X00076512}, number={4}, journal={Behavioral and Brain Sciences}, author={Premack, David and Woodruff, Guy}, year={1978}, pages={515–526}}

@article{BARONCOHEN198537,
title = {Does the autistic child have a “theory of mind” ?},
journal = {Cognition},
volume = {21},
number = {1},
pages = {37-46},
year = {1985},
issn = {0010-0277},
doi = {https://doi.org/10.1016/0010-0277(85)90022-8},
url ={https://www.sciencedirect.com/science/article/pii/0010027785900228},
author = {Simon Baron-Cohen and Alan M. Leslie and Uta Frith}
}

@article{kosinski2024evaluating,
  title={Evaluating large language models in theory of mind tasks},
  author={Kosinski, Michal},
  journal={Proceedings of the National Academy of Sciences},
  volume={121},
  number={45},
  pages={e2405460121},
  year={2024},
  publisher={National Academy of Sciences}
}

@article{ullman2023large,
  title={Large language models fail on trivial alterations to theory-of-mind tasks},
  author={Ullman, Tomer},
  journal={arXiv preprint arXiv:2302.08399},
  year={2023}
}

@inproceedings{shapira2024clever,
  title={Clever hans or neural theory of mind? stress testing social reasoning in large language models},
  author={Shapira, Natalie and Levy, Mosh and Alavi, Seyed Hossein and Zhou, Xuhui and Choi, Yejin and Goldberg, Yoav and Sap, Maarten and Shwartz, Vered},
  booktitle={Proceedings of the 18th Conference of the European Chapter of the Association for Computational Linguistics (Volume 1: Long Papers)},
  pages={2257--2273},
  year={2024}
}

@inproceedings{sap2022neural,
  title={Neural theory-of-mind? on the limits of social intelligence in large lms},
  author={Sap, Maarten and Le Bras, Ronan and Fried, Daniel and Choi, Yejin},
  booktitle={Proceedings of the 2022 conference on empirical methods in natural language processing},
  pages={3762--3780},
  year={2022}
}

@inproceedings{
gu2026simpletom,
title={SimpleToM: Exposing the Gap between Explicit ToM Inference and Implicit ToM Application in {LLM}s},
author={Yuling Gu and Oyvind Tafjord and Hyunwoo Kim and Jared Moore and Ronan Le Bras and Peter Clark and Yejin Choi},
booktitle={The Fourteenth International Conference on Learning Representations},
year={2026},
url={https://openreview.net/forum?id=iE2JmbRJow}
}

@inproceedings{
feng2026inplace,
title={In-Place Test-Time Training},
author={Guhao Feng and Shengjie Luo and Kai Hua and Ge Zhang and Wenhao Huang and Di He and Tianle Cai},
booktitle={The Fourteenth International Conference on Learning Representations},
year={2026},
url={https://openreview.net/forum?id=dTWfCLSoyl}
}

@ARTICLE{2026arXiv260116175Y,
       author = {{Yuksekgonul}, Mert and {Koceja}, Daniel and {Li}, Xinhao and {Bianchi}, Federico and {McCaleb}, Jed and {Wang}, Xiaolong and {Kautz}, Jan and {Choi}, Yejin and {Zou}, James and {Guestrin}, Carlos and {Sun}, Yu},
        title = "{Learning to Discover at Test Time}",
      journal = {arXiv e-prints},
         year = 2026,
        month = jan,
          eid = {arXiv:2601.16175},
        pages = {arXiv:2601.16175},
          doi = {10.48550/arXiv.2601.16175},
archivePrefix = {arXiv},
       eprint = {2601.16175},
 primaryClass = {stat.ML},
       adsurl = {https://ui.adsabs.harvard.edu/abs/2026arXiv260116175Y}
}

@inproceedings{wilf2024think,
  title={Think twice: Perspective-taking improves large language models’ theory-of-mind capabilities},
  author={Wilf, Alex and Lee, Sihyun and Liang, Paul Pu and Morency, Louis-Philippe},
  booktitle={Proceedings of the 62nd Annual Meeting of the Association for Computational Linguistics (Volume 1: Long Papers)},
  pages={8292--8308},
  year={2024}
}

@inproceedings{sclar-etal-2023-minding,
    title = "Minding Language Models' (Lack of) Theory of Mind: A Plug-and-Play Multi-Character Belief Tracker",
    author = "Sclar, Melanie  and
      Kumar, Sachin  and
      West, Peter  and
      Suhr, Alane  and
      Choi, Yejin  and
      Tsvetkov, Yulia",
    editor = "Rogers, Anna  and
      Boyd-Graber, Jordan  and
      Okazaki, Naoaki",
    booktitle = "Proceedings of the 61st Annual Meeting of the Association for Computational Linguistics (Volume 1: Long Papers)",
    month = jul,
    year = "2023",
    address = "Toronto, Canada",
    publisher = "Association for Computational Linguistics",
    url = "https://aclanthology.org/2023.acl-long.780/",
    doi = "10.18653/v1/2023.acl-long.780",
    pages = "13960--13980"
}

@inproceedings{xiao-etal-2025-towards,
    title = "Towards Dynamic Theory of Mind: Evaluating {LLM} Adaptation to Temporal Evolution of Human States",
    author = "Xiao, Yang  and
      Wang, Jiashuo  and
      Xu, Qiancheng  and
      Song, Changhe  and
      Xu, Chunpu  and
      Cheng, Yi  and
      Li, Wenjie  and
      Liu, Pengfei",
    editor = "Che, Wanxiang  and
      Nabende, Joyce  and
      Shutova, Ekaterina  and
      Pilehvar, Mohammad Taher",
    booktitle = "Proceedings of the 63rd Annual Meeting of the Association for Computational Linguistics (Volume 1: Long Papers)",
    month = jul,
    year = "2025",
    address = "Vienna, Austria",
    publisher = "Association for Computational Linguistics",
    url = "https://aclanthology.org/2025.acl-long.1171/",
    doi = "10.18653/v1/2025.acl-long.1171",
    pages = "24036--24057",
    ISBN = "979-8-89176-251-0"
}

@inproceedings{wu-etal-2023-hi,
    title = "Hi-{T}o{M}: A Benchmark for Evaluating Higher-Order Theory of Mind Reasoning in Large Language Models",
    author = "Wu, Yufan  and
      He, Yinghui  and
      Jia, Yilin  and
      Mihalcea, Rada  and
      Chen, Yulong  and
      Deng, Naihao",
    editor = "Bouamor, Houda  and
      Pino, Juan  and
      Bali, Kalika",
    booktitle = "Findings of the Association for Computational Linguistics: EMNLP 2023",
    month = dec,
    year = "2023",
    address = "Singapore",
    publisher = "Association for Computational Linguistics",
    url = "https://aclanthology.org/2023.findings-emnlp.717/",
    doi = "10.18653/v1/2023.findings-emnlp.717",
    pages = "10691--10706"
}

@inproceedings{xu-etal-2024-opentom,
    title = "{O}pen{T}o{M}: A Comprehensive Benchmark for Evaluating Theory-of-Mind Reasoning Capabilities of Large Language Models",
    author = "Xu, Hainiu  and
      Zhao, Runcong  and
      Zhu, Lixing  and
      Du, Jinhua  and
      He, Yulan",
    editor = "Ku, Lun-Wei  and
      Martins, Andre  and
      Srikumar, Vivek",
    booktitle = "Proceedings of the 62nd Annual Meeting of the Association for Computational Linguistics (Volume 1: Long Papers)",
    month = aug,
    year = "2024",
    address = "Bangkok, Thailand",
    publisher = "Association for Computational Linguistics",
    url = "https://aclanthology.org/2024.acl-long.466/",
    doi = "10.18653/v1/2024.acl-long.466",
    pages = "8593--8623"
}

@inproceedings{hu2022lora,
  title={LoRA: Low-Rank Adaptation of Large Language Models},
  author={Hu, Edward J. and Shen, Yelong and Wallis, Phillip and Allen-Zhu, Zeyuan and Li, Yuanzhi and Wang, Shean and Wang, Liang and Chen, Weizhu},
  booktitle={Proceedings of the 10th International Conference on Learning Representations (ICLR)},
  year={2022},
  url={https://openreview.net/forum?id=H1eA7AEtvS}
}

@inproceedings{
loshchilov2018decoupled,
title={Decoupled Weight Decay Regularization},
author={Ilya Loshchilov and Frank Hutter},
booktitle={International Conference on Learning Representations},
year={2019},
url={https://openreview.net/forum?id=Bkg6RiCqY7},
}

@misc{qwen2.5,
    title = {Qwen2.5: A Party of Foundation Models},
    url = {https://qwenlm.github.io/blog/qwen2.5/},
    author = {Qwen Team},
    month = {September},
    year = {2024}
}

@article{grattafiori2024llama,
  title={The llama 3 herd of models},
  author={Grattafiori, Aaron and Dubey, Abhimanyu and Jauhri, Abhinav and Pandey, Abhinav and Kadian, Abhishek and Al-Dahle, Ahmad and Letman, Aiesha and Mathur, Akhil and Schelten, Alan and Vaughan, Alex and others},
  journal={arXiv preprint arXiv:2407.21783},
  year={2024}
}

@misc{openai2025gptoss120bgptoss20bmodel,
      title={gpt-oss-120b \& gpt-oss-20b Model Card}, 
      author={OpenAI},
      year={2025},
      eprint={2508.10925},
      archivePrefix={arXiv},
      primaryClass={cs.CL},
      url={https://arxiv.org/abs/2508.10925}, 
}

@inproceedings{Sun2019TestTimeTW,
  title={Test-Time Training with Self-Supervision for Generalization under Distribution Shifts},
  author={Yu Sun and X. Wang and Zhuang Liu and John Miller and Alexei A. Efros and Moritz Hardt},
  booktitle={International Conference on Machine Learning},
  year={2019}
}

@inproceedings{NEURIPS2021_b618c321,
 author = {Liu, Yuejiang and Kothari, Parth and van Delft, Bastien and Bellot-Gurlet, Baptiste and Mordan, Taylor and Alahi, Alexandre},
 booktitle = {Advances in Neural Information Processing Systems},
 editor = {M. Ranzato and A. Beygelzimer and Y. Dauphin and P.S. Liang and J. Wortman Vaughan},
 pages = {21808--21820},
 publisher = {Curran Associates, Inc.},
 title = {TTT++: When Does Self-Supervised Test-Time Training Fail or Thrive?},
 url = {https://proceedings.neurips.cc/paper_files/paper/2021/file/b618c3210e934362ac261db280128c22-Paper.pdf},
 volume = {34},
 year = {2021}
}

@article{moghaddam2023boosting,
  title={Boosting theory-of-mind performance in large language models via prompting},
  author={Moghaddam, Shima Rahimi and Honey, Christopher J},
  journal={arXiv preprint arXiv:2304.11490},
  year={2023}
}

@inproceedings{10.5555/3600270.3602070, author = {Wei, Jason and Wang, Xuezhi and Schuurmans, Dale and Bosma, Maarten and Ichter, Brian and Xia, Fei and Chi, Ed H. and Le, Quoc V. and Zhou, Denny}, title = {Chain-of-thought prompting elicits reasoning in large language models}, year = {2022}, isbn = {9781713871088}, publisher = {Curran Associates Inc.}, address = {Red Hook, NY, USA}, booktitle = {Proceedings of the 36th International Conference on Neural Information Processing Systems}, articleno = {1800}, numpages = {14}, location = {New Orleans, LA, USA}, series = {NIPS '22} }

@article{https://doi.org/10.1111/j.1468-0017.1986.tb00324.x,
author = {GORDON, ROBERT M.},
title = {Folk Psychology as Simulation},
journal = {Mind \& Language},
volume = {1},
number = {2},
pages = {158-171},
doi = {https://doi.org/10.1111/j.1468-0017.1986.tb00324.x},
url = {https://onlinelibrary.wiley.com/doi/abs/10.1111/j.1468-0017.1986.tb00324.x},
eprint = {https://onlinelibrary.wiley.com/doi/pdf/10.1111/j.1468-0017.1986.tb00324.x},
year = {1986}
}

@article{wimmer1983beliefs,
  title={Beliefs about beliefs: Representation and constraining function of wrong beliefs in young children's understanding of deception},
  author={Wimmer, Heinz and Perner, Josef},
  journal={Cognition},
  volume={13},
  number={1},
  pages={103--128},
  year={1983},
  publisher={Elsevier}
}

@article{quesque2020theory,
  title={What do theory-of-mind tasks actually measure? Theory and practice},
  author={Quesque, Fran{\c{c}}ois and Rossetti, Yves},
  journal={Perspectives on psychological science},
  volume={15},
  number={2},
  pages={384--396},
  year={2020},
  publisher={Sage Publications Sage CA: Los Angeles, CA}
}

@incollection{hiebert2013conceptual,
  title={Conceptual and procedural knowledge in mathematics: An introductory analysis},
  author={Hiebert, James and Lefevre, Patricia},
  booktitle={Conceptual and procedural knowledge},
  pages={1--27},
  year={2013},
  publisher={Routledge}
}

@article{hutchins2016explicit,
  title={Explicit vs. applied theory of mind competence: A comparison of typically developing males, males with ASD, and males with ADHD},
  author={Hutchins, Tiffany L and Prelock, Patricia A and Morris, Hope and Benner, Joy and LaVigne, Timothy and Hoza, Betsy},
  journal={Research in Autism spectrum disorders},
  volume={21},
  pages={94--108},
  year={2016},
  publisher={Elsevier}
}

@article{apperly2012theory,
  title={What is “theory of mind”? Concepts, cognitive processes and individual differences},
  author={Apperly, Ian A},
  journal={Quarterly journal of experimental psychology},
  volume={65},
  number={5},
  pages={825--839},
  year={2012},
  publisher={SAGE Publications Sage UK: London, England}
}

@article{trott2020comprehenders,
  title={When do comprehenders mentalize for pragmatic inference?},
  author={Trott, Sean and Bergen, Benjamin},
  journal={Discourse Processes},
  volume={57},
  number={10},
  pages={900--920},
  year={2020},
  publisher={Taylor \& Francis}
}

@inproceedings{saxena-etal-2025-streamlining,
    title = "Streamlining {LLM}s: Adaptive Knowledge Distillation for Tailored Language Models",
    author = "Saxena, Prajvi  and
      Janzen, Sabine  and
      Maa{\ss}, Wolfgang",
    editor = "Ebrahimi, Abteen  and
      Haider, Samar  and
      Liu, Emmy  and
      Haider, Sammar  and
      Leonor Pacheco, Maria  and
      Wein, Shira",
    booktitle = "Proceedings of the 2025 Conference of the Nations of the Americas Chapter of the Association for Computational Linguistics: Human Language Technologies (Volume 4: Student Research Workshop)",
    month = apr,
    year = "2025",
    address = "Albuquerque, USA",
    publisher = "Association for Computational Linguistics",
    url = "https://aclanthology.org/2025.naacl-srw.43/",
    doi = "10.18653/v1/2025.naacl-srw.43",
    pages = "448--455",
    ISBN = "979-8-89176-192-6"
}

\appendix

\section{Background on Theory of Mind}\label{background}
\subsection{What is Theory of Mind?}
Theory of Mind (ToM) refers to the capacity to attribute mental states, beliefs, intentions, desires, emotions, knowledge to other agents, and to recognize that those states may differ from one's own and from the actual state of the world \citep{Premack_Woodruff_1978}. The term was introduced in the context of primate cognition and subsequently became central to developmental psychology, where the canonical test is the false belief task: a child must predict what an agent will do based on a belief the agent holds that the child knows to be false \citep{BARONCOHEN198537, wimmer1983beliefs}.
Two criteria, formalized by \citet{quesque2020theory}, define tasks that genuinely validate ToM rather than measuring surface heuristics: (1) the task must require distinguishing another agent's mental state from one's own knowledge of the world; (2) lower-level processes such as associative learning or keyword matching must be ruled out as explanations for correct performance. Both criteria bear directly on how we evaluate LLMs, since models trained on large corpora may answer ToM questions correctly by pattern matching to similar training examples rather than by genuine mental-state inference.

\subsection{Explicit vs. Applied ToM: Grounding in Psychology}
The distinction between explicit and applied ToM has roots in broader cognitive theories contrasting declarative and procedural knowledge \citep{hiebert2013conceptual}, and in clinical and developmental psychology literature on ToM component processes \citep{hutchins2016explicit, gu2026simpletom}. Explicit ToM is knowing that another agent holds a mental state, that is a representational capacity. Applied ToM uses that knowledge to predict behavior, anticipate consequences, or evaluate actions, which is a deployment capacity. The two can dissociate: individuals with certain clinical profiles demonstrate intact explicit ToM (correctly identifying others' beliefs) alongside impaired applied ToM (failing to use those beliefs in social prediction) \citep{apperly2012theory}. This maps onto a distinction \citet{trott2020comprehenders} draw between sampling mental state information and deploying it for pragmatic inference, and onto \citet{apperly2012theory} framework of inference, storage, and use. In our operationalization, explicit questions $\mathcal{B}_\mathrm{E}$ assess mental-state identification, while applied questions $\mathcal{B}_\mathrm{A}$ assess the use of mental-state information to predict behavior
or evaluate actions.

\subsection{Why the Gap Matters Beyond Benchmarks}
The practical importance of applied ToM scales with the autonomy and social embeddedness of AI systems. Three deployment contexts illustrate the stakes: Assistive AI. A personal assistant that recognizes a user is unaware of a risk but fails to predict that the user will therefore take the risky action cannot intervene appropriately. The recognition is necessary but not sufficient; prediction is the action-relevant step.
Evaluative AI. Systems used to assess the reasonableness of human actions in legal, medical, or institutional contexts must evaluate actions against the mental state of the actor at the time, not against the full information available to the evaluating system. Failing to apply a recognized false belief to a judgment of reasonableness produces systematically biased evaluations, penalizing actors for outcomes they could not have anticipated. Conversational AI. A dialogue system that models a user's beliefs about a topic can provide genuinely personalized responses not just statistically likely ones. Applied ToM is the mechanism that converts a user model (explicit) into a communication strategy (applied).

\subsection{Connection to Classical ToM Tests}
The benchmarks used in this paper extend classical false-belief tasks in several directions. SimpleToM removes explicit mentalizing cues (words like "sees," "believes," "thinks") that allow keyword-based success, requiring commonsense inference about information asymmetry which is closer to how ToM operates in naturalistic settings. Hi-ToM extends to higher-order recursive beliefs (what Alice thinks Bob thinks), which have no direct analogue in classical developmental tasks but are common in adult social cognition. DynToM adds temporal evolution of belief states, reflecting the dynamic character of real social interaction. OpenToM introduces multi-agent narratives with heterogeneous information access. Together these benchmarks span the space from simple false-belief tracking to temporally extended social inference, allowing us to ask not just whether models have ToM but where, structurally, the explicit-to-applied gap emerges and whether it is recoverable.

%% ============================================================
%% APPENDIX B: BENCHMARK DESCRIPTIONS
%% ============================================================
\section{Benchmark Descriptions}
\label{app:benchmark_descriptions}

This section provides descriptions of the four ToM benchmarks used in our evaluation. Each description covers the benchmark's design rationale, question taxonomy, and the specific subset we evaluated.

\paragraph{Support--query separation.}
We verified that held-out queries are excluded from
their support sets and that query reference labels
are not supplied during prompting or adaptation.
Answered supports may nevertheless provide facts
relevant to the query. We therefore measure recovery
under this assistance; matched ICL and BSTTT compare
how the same answered supports are used.

\subsection{SimpleToM}
\label{app:bench_simpletom}

SimpleToM~\citep{gu2026simpletom} probes whether LLMs can move beyond \emph{recognizing} an agent's mental state to \emph{applying} that state for behavior prediction and judgment.
Each item is a concise two-sentence story grounded in an everyday scenario involving information asymmetry (e.g., hidden defects in grocery items, undisclosed conflicts of interest, or unobserved unethical actions). Crucially, the stories avoid explicit mentalizing verbs such as ``sees,'' ``thinks,'' or ``believes''; instead, the character's unawareness must be inferred through commonsense reasoning about the situation.

Each story generates a matched triplet of binary-choice questions:

\begin{enumerate}[leftmargin=2em,itemsep=2pt]
    \item \textbf{Mental-state (explicit):} ``Is \texttt{<Person X>} likely to be aware that \texttt{<key information>}?'' Tests whether the model can identify the character's information state.
    \item \textbf{Behavior prediction (applied):} ``What will \texttt{<Person X>} likely do next?'' with two options, the ``unaware'' action (what the character would do given their false belief) and the ``aware'' action (what they would do if they knew the hidden information).
    \item \textbf{Judgment (applied):} Specifies that the character performed the ``unaware'' action and asks whether this was reasonable. Requires integrating the inferred mental state with evaluative commonsense.
\end{enumerate}

The dataset contains 1,147 human-validated stories (3,441 questions) across ten information-asymmetry scenarios.
All questions are binary ($C{=}2$, chance~$= 50\%$).
The original paper~\citep{gu2026simpletom} reports that frontier models achieve $>$95\% on mental-state questions but drop to below 50\% on behavior prediction and below 25\% on judgment, a striking explicit-to-applied gap that directly motivates our work.

\paragraph{Our operationalization.}
We use the mental-state question as $\mathcal{B}_\mathrm{E}$ and the behavior and judgment questions as $\mathcal{B}_\mathrm{A}$. Each episode uses one answered behavior question from the family as support for a held-out query. We verified that each held-out query is excluded from its support set and that its reference label is not used during prompting or adaptation.

\subsection{Hi-ToM}
\label{app:bench_hitom}

Hi-ToM~\citep{wu-etal-2023-hi} evaluates higher-order recursive Theory of Mind reasoning. Each story describes a sequence of events involving multiple agents, objects, and locations, with interactions including communication, witnessed events, unwitnessed events, and both public and private deceptive claims.

Each story is paired with five questions targeting orders 0 through 4 of ToM reasoning:
\begin{itemize}[leftmargin=2em,itemsep=2pt]
    \item \textbf{0th-order:} Factual world-state queries (``Where is the object?'').
    \item \textbf{1st-order:} ``Where does Agent~A think the object is?''
    \item \textbf{2nd-order:} ``Where does Agent~A think Agent~B thinks the object is?''
    \item \textbf{3rd/4th-order:} Deeper recursive nesting.
\end{itemize}

Each question has 15 answer choices ($C{=}15$, chance~$\approx 6.7\%$).
Performance degrades systematically with order, and the inclusion of deceptive claims makes higher-order questions particularly challenging.

\paragraph{Our operationalization.}
Hi-ToM does not provide a paired recognition--application decomposition. We therefore report overall accuracy and intervention gains without computing an explicit-to-applied gap.

\subsection{OpenToM}
\label{app:bench_opentom}

OpenToM~\citep{xu-etal-2024-opentom} is a comprehensive ToM benchmark featuring multi-agent narratives with rich social interactions. Stories involve characters with varying degrees of access to information, and questions probe both attitude inference and multihop first-order reasoning.

Our evaluation uses two question families:
\begin{itemize}[leftmargin=2em,itemsep=2pt]
    \item \textbf{Attitude questions (explicit):} Probe a character's sentiment, preference, or emotional state toward an entity.
    \item \textbf{Multihop first-order questions (applied):} Require chaining information across multiple events and characters to arrive at a first-order belief attribution.
\end{itemize}

Answer choices vary by question type ($C \in \{3, \ldots, 6\}$, chance~$\approx 25\%$).

\paragraph{Our operationalization.}
Attitude questions serve as $\mathcal{B}_\mathrm{E}$; multihop questions serve as $\mathcal{B}_\mathrm{A}$.
For the episodes, questions from the pooled narrative serve as support ($|\mathcal{S}| \leq 4$).

\subsection{DynToM}
\label{app:bench_dyntom}

DynToM~\citep{xiao-etal-2025-towards} extends ToM evaluation to the temporal domain.    Unlike static benchmarks where mental states are fixed, DynToM requires tracking how beliefs, emotions, intentions, and actions \emph{evolve} across a sequence of five temporally ordered social scenarios.

Each trial consists of:
\begin{itemize}[leftmargin=2em,itemsep=2pt]
    \item \textbf{Character descriptions:} Background information about the two characters involved.
    \item \textbf{Five scenarios:} Temporally ordered events that progressively change the characters' mental states.
    \item \textbf{Understanding questions (\texttt{type\_a}, explicit):} Probe the character's current belief, emotion, intention, or action at a specific scenario. These are local state-tracking questions.
    \item \textbf{Transformation questions (\texttt{type\_c}, \texttt{type\_d}, applied):} Require reasoning about how mental states \emph{shift} across scenarios e.g., ``How did Character~A's belief change from Scenario~2 to Scenario~4?''
\end{itemize}

Each question has 7--8 answer options ($C \in \{7, 8\}$, chance~$\approx 14\%$). The benchmark comprises 1,100 social contexts, 5,500 scenarios, and 78,100 questions.
State-of-the-art LLMs underperform humans by 44.7\% on average~\citep{xiao-etal-2025-towards}.

\paragraph{Our operationalization.}
Understanding questions (\texttt{type\_a}) serve as $\mathcal{B}_\mathrm{E}$; transformation questions (\texttt{type\_c}, \texttt{type\_d}) serve as $\mathcal{B}_\mathrm{A}$.
For BSTTT episodes, the \texttt{type\_a} questions among the pooled trial serve as support.

%% ============================================================
%% APPENDIX C: DETAILED METHOD DESCRIPTIONS AND PROMPTS
%% ==========================================================
\section{Baseline Methods and Prompt Templates}
\label{app:method_details}

This appendix describes the evaluated methods and their
prompt formats. We cover direct scoring, structured
prompting, matched ICL, and BSTTT with its prompting
compositions. Table~\ref{tab:appendix_method_map} maps
these methods to their benchmark-specific implementations.

\begin{table*}[t]
    \caption{\textbf{Benchmark-specific method map.}
    Fixed-weight methods modify the input without updating
    model parameters. Adaptation methods use BSTTT alone
    or in combination with prompting. Scratchpad-Oracle
    supplies annotated mental states as a separate control.}
    \label{tab:appendix_method_map}
    \centering
    \footnotesize
    \begin{tabular}{@{}lp{5.0cm}p{6.0cm}@{}}
        \toprule
        \textbf{Benchmark}
        & \textbf{Fixed-weight methods and controls}
        & \textbf{Adaptation and composed methods} \\
        \midrule
        SimpleToM
        & Vanilla, CoT, MS-Reminder, matched ICL
        & BSTTT-NTL, BSTTT-AR, MS-Reminder+BSTTT \\
        Hi-ToM / OpenToM
        & Vanilla, CoT, Scratchpad, SimToM,
          SymbolicToM, matched ICL
        & BSTTT-NTL, BSTTT-AR, Scratchpad+BSTTT,
          SimToM+BSTTT, SymbolicToM+BSTTT \\
        DynToM
        & Frozen, CoT, Scratchpad-Frozen,
          matched ICL, Scratchpad-Oracle
        & BSTTT-NTL, BSTTT-AR, Scratchpad-TTT,
          Hierarchical-TTT \\
        \bottomrule
    \end{tabular}
\end{table*}

% This appendix provides detailed descriptions of each method evaluated in the paper, including its exact prompt templates used for our implementation.
% We organize methods into three categories: (1)~frozen baselines with no prompt scaffold, (2)~structured prompting methods from the ToM literature, and (3)~our diagnostic adaptation methods.
% Exact runnable prompt files are included in the code release.

% \begin{table*}[t]
%     \caption{\textbf{Benchmark-specific method map.}
%     The main-text method families instantiate differently across benchmarks; this table shows which concrete prompts and controls are used where.}
%     \label{tab:appendix_method_map}
%     \footnotesize
%     \begin{tabular}{@{}lp{5.0cm}p{6.0cm}@{}}
%         \toprule
%         \textbf{Benchmark} & \textbf{Prompt-only methods} & \textbf{Adaptation / hybrid methods} \\
%         \midrule
%         SimpleToM & Vanilla, CoT, MS-Reminder & BSTTT-NTL, BSTTT-AR, MS-Reminder+BSTTT \\
%         Hi-ToM / OpenToM & Vanilla, CoT, Scratchpad, SimToM, SymbolicToM & BSTTT-NTL, BSTTT-AR, Scratchpad+BSTTT, SimToM+BSTTT, SymbolicToM+BSTTT \\
%         DynToM & Frozen, CoT, Scratchpad-Frozen & BSTTT, Scratchpad-Oracle, Scratchpad-TTT, Hierarchical-TTT \\
%         \bottomrule
%     \end{tabular}
% \end{table*}

%% --- C.1  Frozen Baselines ---
\subsection{Frozen Baselines}
\label{app:frozen_baselines}

\paragraph{Frozen / Vanilla.}
The frozen baseline performs direct multiple-choice scoring: each answer option is appended to the prompt, the model computes the log-likelihood of each option string, and the highest-scoring option is selected.
No reasoning scaffold or chain-of-thought instruction is provided.

\begin{tcolorbox}[colback=gray!5, colframe=gray!40, boxrule=0.3pt, left=4pt, right=4pt, top=3pt, bottom=3pt, title={\small\textbf{Vanilla Prompt Template}}]
\small\ttfamily
Story:\\
\mbox{[story]}\\[2pt]
Question:\\
\mbox{[question]}\\[2pt]
Choices:\\
A. [choice 1]\\
B. [choice 2]\\[2pt]
Answer:
\end{tcolorbox}

\paragraph{Chain-of-Thought (CoT).}
The CoT baseline adds an explicit step-by-step reasoning instruction before option scoring. This follows the standard CoT prompting paradigm~\citep{10.5555/3600270.3602070}, which has been shown to improve explicit ToM reasoning but with inconsistent effects on applied prediction~\citep{moghaddam2023boosting}.

\begin{tcolorbox}[colback=gray!5, colframe=gray!40, boxrule=0.3pt, left=4pt, right=4pt, top=3pt, bottom=3pt, title={\small\textbf{CoT Prompt Template}}]
\small\ttfamily
Story:\\
\mbox{[story]}\\[2pt]
Question:\\
\mbox{[question]}\\[2pt]
Let's think step by step to reason about this question, then give your final answer.\\[2pt]
Choices:\\
A. [choice 1]\\
B. [choice 2]\\[2pt]
Answer (state the letter or choice):
\end{tcolorbox}

\subsection{Matched In-Context Learning}
\label{app:matched_icl}

Matched ICL presents answered support questions in the
query prompt without updating model parameters. Support selection follows the benchmark-specific episode construction in Appendix~\ref{app:benchmark_descriptions}. Each demonstration contains a support question, its
candidate answers, and its correct answer. Query options
are scored by conditional log-likelihood, as in the
frozen baseline. This comparison examines recovery
through contextual demonstrations alongside recovery
through BSTTT adaptation.

For SimpleToM, each demonstration includes the story.
Hi-ToM and OpenToM present the shared narrative before
the support questions; DynToM supplies the character
descriptions and ordered trial scenarios.

\begin{tcolorbox}[
    colback=gray!5, colframe=gray!40, boxrule=0.3pt,
    left=4pt, right=4pt, top=3pt, bottom=3pt,
    title={\small\textbf{Matched ICL: SimpleToM Prompt Template}}
]
\small\ttfamily
Use the answered examples from this scenario family
to answer the final question.\\[2pt]
Answered support example 1:\\
Story:\\
\mbox{[support story]}\\[2pt]
Question:\\
\mbox{[support question]}\\[2pt]
Choices:\\
A. [support choice 1]\\
B. [support choice 2]\\[2pt]
Correct answer: [support answer]\\[2pt]
Final story:\\
\mbox{[query story]}\\[2pt]
Final question:\\
\mbox{[query question]}\\[2pt]
Choices:\\
A. [query choice 1]\\
B. [query choice 2]\\[2pt]
Answer:
\end{tcolorbox}

%% --- C.2  SimpleToM-Specific Methods ---
\subsection{SimpleToM-Specific Methods}
\label{app:simpletom_methods}

\paragraph{Mental-State Reminder (MS-Reminder).}
This intervention was introduced and studied in the original SimpleToM paper~\citep{gu2026simpletom} as the ``MS remind'' condition. The key idea is to close the explicit-to-applied gap by patching the model's own mental-state inference into the prompt when answering behavior or judgment questions.

The procedure has two steps:
\begin{enumerate}[leftmargin=2em,itemsep=2pt]
    \item \textbf{Predict the mental state:} The mental-state question for the same scenario is answered using the vanilla prompt. The model's predicted answer (e.g., ``No'') is recorded.
    \item \textbf{Inject the reminder:} When scoring the behavior or judgment question, the mental-state question and the model's predicted answer are prepended to the prompt as a one-line conversation history.
\end{enumerate}

\citet{gu2026simpletom} showed that this intervention boosts behavior prediction accuracy substantially (e.g., from 49.5\% to 82.8\% for GPT-4o) but has more modest effects on judgment. These results show that the benefits of mental-state reminders vary across applied question types.

\begin{tcolorbox}[colback=gray!5, colframe=gray!40, boxrule=0.3pt, left=4pt, right=4pt, top=3pt, bottom=3pt, title={\small\textbf{MS-Reminder Prompt Template} (from~\citet{gu2026simpletom})}]
\small\ttfamily
Story:\\
\mbox{[story]}\\[2pt]
Question: Is [Person X] likely to be aware that ``[key information]''? \\
(A) No (B) Yes \\
Answer: [model-predicted mental-state answer]\\[2pt]
Question: [behavior or judgment question]\\
(A) [choice 1] (B) [choice 2]\\[2pt]
What is the correct answer? Respond with just ``(A)'' or ``(B)''
\end{tcolorbox}

\paragraph{MS-Reminder+BSTTT.}
The hybrid SimpleToM condition combines the mental-state reminder with episodic AR adaptation.
Both support questions and the held-out query use the reminder-style prompt whenever the query target is behavior or judgment.
This tests whether explicit input-level state reminders and weight-level adaptation are complementary on the cleanest explicit-to-applied split.

%% --- C.3  External ToM Prompting Baselines ---
\subsection{SimToM: Perspective-Taking via Simulation Theory}
\label{app:simtom}

SimToM~\citep{wilf2024think} is a two-stage prompting framework inspired by Simulation Theory from cognitive science~\citep{https://doi.org/10.1111/j.1468-0017.1986.tb00324.x}.
The core insight is that ToM failures in LLMs often stem from the model reasoning from its omniscient perspective rather than from the character's limited perspective. SimToM addresses this by explicitly decomposing ToM inference into perspective-taking and question-answering.

\paragraph{Stage~1: Perspective-Taking.}
The model is prompted to identify which events in the story are known to the queried character.
This step filters the narrative to only the information accessible from that character's viewpoint, effectively simulating their epistemic state.
The prompt follows the template from~\citet{wilf2024think}:

\begin{tcolorbox}[colback=blue!3, colframe=blue!30, boxrule=0.3pt, left=4pt, right=4pt, top=3pt, bottom=3pt, title={\small\textbf{SimToM Stage~1: Perspective-Taking} (adapted from~\citet{wilf2024think})}]
\small\ttfamily
You are performing perspective-taking for Theory of Mind.\\
Task: extract only the facts that relevant agents can observe and infer.\\
Do not answer the final question yet.\\[2pt]
Story:\\
\mbox{[story]}\\[2pt]
Question:\\
\mbox{[question]}\\[2pt]
Output format:\\
1) Relevant agents\\
2) Observable events per agent\\
3) Inferred beliefs per agent
\end{tcolorbox}

The model generates free-form perspective notes (up to 192 tokens, temperature~0).

\paragraph{Stage~2: Question-Answering from Filtered Perspective.}
The generated perspective notes are prepended to the
multiple-choice prompt, which retains the original
narrative and instructs the model to answer using
the perspective-filtered notes:

\begin{tcolorbox}[colback=blue!3, colframe=blue!30, boxrule=0.3pt, left=4pt, right=4pt, top=3pt, bottom=3pt, title={\small\textbf{SimToM Stage~2: Filtered QA}}]
\small\ttfamily
Perspective-Taking Notes:\\
\mbox{[stage 1 output]}\\[2pt]
Use only these perspective-filtered notes to answer.\\[2pt]
Story: [story]\\
Question: [question]\\
Choices: [options]\\
Answer:
\end{tcolorbox}

% \paragraph{Key properties.}
% SimToM is training-free, requires no additional data, and composes naturally with BSTTT because the perspective-taking step modifies the input while BSTTT modifies the weights.
% \citet{wilf2024think} demonstrated that SimToM substantially outperforms standard zero-shot and CoT prompting on several ToM benchmarks.

\subsection{SymbolicToM: Symbolic Belief-State Tracking}
\label{app:symbolictom}

SymbolicToM~\citep{sclar-etal-2023-minding} takes a fundamentally different approach from SimToM: rather than filtering the narrative to a character's perspective, it constructs an explicit \emph{symbolic belief graph} that tracks each entity's beliefs, their estimations of other entities' beliefs, and higher-order nesting, all through graphical representations.

The original SymbolicToM system~\citep{sclar-etal-2023-minding} is a multi-step decoding-time algorithm that:
\begin{enumerate}[leftmargin=2em,itemsep=2pt]
    \item Parses the story sentence-by-sentence, extracting state updates as structured triples.
    \item Maintains a belief graph where nodes represent entities and edges represent belief relations (including nested beliefs for higher-order reasoning).
    \item Retrieves the relevant subgraph for the queried entity and reasoning depth.
    \item Answers the question conditioned on the retrieved symbolic state.
\end{enumerate}

\paragraph{Our operationalization.}
We implement SymbolicToM as a two-stage prompting baseline that captures the core logic of the original system by building an explicit belief tracker and then answering from that tracker, while remaining compatible with our log-likelihood scoring pipeline.

\begin{tcolorbox}[colback=green!3, colframe=green!30, boxrule=0.3pt, left=4pt, right=4pt, top=3pt, bottom=3pt, title={\small\textbf{SymbolicToM Stage~1: Belief Tracker Construction} (operationalizing~\citet{sclar-etal-2023-minding})}]
\small\ttfamily
Build a symbolic belief tracker for this story.\\
Represent each state update as:\\
{[time]} agent | belief | evidence\\
Include nested beliefs if present.\\
Do not answer the final question yet.\\[2pt]
Story:\\
\mbox{[story]}\\[2pt]
Question:\\
\mbox{[question]}\\[2pt]
Belief Tracker:
\end{tcolorbox}

\begin{tcolorbox}[colback=green!3, colframe=green!30, boxrule=0.3pt, left=4pt, right=4pt, top=3pt, bottom=3pt, title={\small\textbf{SymbolicToM Stage~2: Tracker-Conditioned Answering}}]
\small\ttfamily
Belief Tracker State:\\[2pt]
[tracker output from Stage~1]\\[2pt]
Answer by following the tracker transitions and constraints.\\[2pt]
Story: [story]\\
Question: [question]\\
Choices: [options]\\
Answer:
\end{tcolorbox}

\paragraph{Comparison with SimToM.}
While SimToM filters information by perspective and SymbolicToM builds explicit state representations, both methods externalize latent variables that standard prompting leaves implicit. Their relative effectiveness varies across models (Table~\ref{tab:main_hitom_datasetwise}).

\subsection{Scratchpad: Structured Belief-State Scaffold}
\label{app:scratchpad}

The Scratchpad baseline prepends a structured mental-state scaffold to the story prompt.
Unlike SimToM and SymbolicToM, it does not require a separate generation pass; instead, it instructs the model to internally organize its reasoning around explicit state variables before answering.
The scaffold identifies the key dimensions of ToM reasoning and directs the model to resolve conflicts between the world state and each agent's belief.

The exact focus instruction is adapted to the question type:
\begin{itemize}[leftmargin=2em,itemsep=2pt]
    \item \textbf{Attitude/emotion questions:} ``Prioritize latent mental states (beliefs, emotions, preferences, intentions).''
    \item \textbf{Multihop/recursive questions:} ``Prioritize nested/recursive beliefs (who thinks what about whom).''
    \item \textbf{Location questions:} ``Track physical state transitions and each agent's belief about locations.''
\end{itemize}

\begin{tcolorbox}[colback=orange!3, colframe=orange!30, boxrule=0.3pt, left=4pt, right=4pt, top=3pt, bottom=3pt, title={\small\textbf{Scratchpad Prompt Template}}]
\small\ttfamily
Mental-State Scratchpad:\\
- Entities: identify key agents and objects.\\
- State variables: belief / emotion / intention / action.\\
- Focus for this question: [question-specific focus].\\
- Resolve conflicts between world state and each agent's belief before answering.\\[2pt]
Story:\\
\mbox{[story]}\\[2pt]
Question:\\
\mbox{[question]}\\[2pt]
Choices:\\
\mbox{[multiple-choice options]}\\[2pt]
Answer:
\end{tcolorbox}

\subsection{Composed Methods: Prompt+BSTTT Hybrids}
\label{app:hybrid_methods}

Scratchpad+BSTTT, SimToM+BSTTT, and SymbolicToM+BSTTT all use the AR adaptation objective on in-scenario support questions. The prompt scaffold is applied at query time after adaptation.
For the two-stage methods (SimToM, SymbolicToM), the stage-1 perspective notes or tracker are generated before adaptation using the frozen model, and the adapted model is then evaluated on the corresponding stage-2 multiple-choice prompt.

This design ensures that the composition is clean: BSTTT modifies the model's internal representations through weight updates, while the prompt scaffold modifies the input structure.
As shown in Eq.~\ref{eq:composition}, the two interventions compose naturally because they operate on orthogonal dimensions.

%% --- C.4  DynToM Methods ---
\subsection{DynToM-Specific Methods}
\label{app:dyntom_methods}

DynToM uses trial-level context rather than a single short narrative.
Each trial contains five temporally ordered scenarios, and the state-tracking questions (\texttt{type\_a}) provide a natural substrate for constructing explicit scratchpads before evaluating the harder transformation questions (\texttt{type\_c}, \texttt{type\_d}).

\paragraph{Shared DynToM query prompt.}
All DynToM conditions use the full trial context: character information, the ordered scenario texts, the query, and the answer choices.
CoT adds a single instruction to reason step by step about beliefs, emotions, intentions, and actions before scoring the options.

\begin{tcolorbox}[colback=gray!5, colframe=gray!40, boxrule=0.3pt, left=4pt, right=4pt, top=3pt, bottom=3pt, title={\small\textbf{DynToM Base Prompt}}]
\small\ttfamily
Characters:\\
\mbox{[character descriptions]}\\[2pt]
[optional scratchpad block]\\[2pt]
Scenario 1: [text]\\
Scenario 2: [text]\\
Scenario 3: [text]\\
Scenario 4: [text]\\
Scenario 5: [text]\\[2pt]
Based on the scenarios above, answer the following multiple-choice question.\\
Question: [question]\\[2pt]
Options:\\
a. [choice 1]\\
b. [choice 2]\\
...\\[2pt]
Answer:
\end{tcolorbox}

\paragraph{Scratchpad-Frozen.}
For transformation questions, we first answer or predict the trial's \texttt{type\_a} state-tracking questions with the frozen model and place those answers into a deterministic $5 \times 4$ table with rows for scenarios 1--5 and columns for belief, emotion, intention, and action.
That table is then prepended to the transformation-question prompt.

\paragraph{Scratchpad-Oracle.}
Scratchpad-Oracle uses the same table format, but fills it with gold \texttt{type\_a} answers rather than model predictions.
This is a diagnostic-only upper-bound control: it asks how much transformation accuracy could improve if local state tracking were perfect.

\paragraph{Scratchpad-TTT.}
Scratchpad-TTT first adapts LoRA weights on the trial-local \texttt{type\_a} questions using AR.
The adapted model is then used to populate the same scratchpad table, and transformation questions are answered with both the explicit scratchpad and the adapted weights.

\paragraph{Hierarchical-TTT.}
Hierarchical-TTT uses the same trial-local AR adaptation on \texttt{type\_a} questions, but it removes the explicit scratchpad table at query time.
Transformation questions are therefore answered directly from the adapted model with the full trial context, isolating the contribution of weight adaptation from the contribution of an externalized state summary.

\begin{tcolorbox}[colback=gray!5, colframe=gray!40, boxrule=0.3pt, left=4pt, right=4pt, top=3pt, bottom=3pt, title={\small\textbf{DynToM Scratchpad Table Format}}]
\small\ttfamily
Mental State Scratchpad\\
Scenario 1: belief=[...]; emotion=[...]; intention=[...]; action=[...]\\
Scenario 2: belief=[...]; emotion=[...]; intention=[...]; action=[...]\\
Scenario 3: belief=[...]; emotion=[...]; intention=[...]; action=[...]\\
Scenario 4: belief=[...]; emotion=[...]; intention=[...]; action=[...]\\
Scenario 5: belief=[...]; emotion=[...]; intention=[...]; action=[...]
\end{tcolorbox}

%% ============================================================
%% APPENDIX D: BSTTT IMPLEMENTATION DETAILS
%% ============================================================
\section{BSTTT Implementation Details}
\label{app:bsttt_details}

\subsection{LoRA Configuration}

All BSTTT experiments use the following fixed configuration unless otherwise noted (\citep{hu2022lora, Sun2019TestTimeTW, saxena-etal-2025-streamlining}):

\begin{table}[h]
    \centering
    \small
    \caption{\textbf{BSTTT hyperparameters.} Fixed across all AR vs.\ NTL comparisons.}
    \label{tab:bsttt_hyperparams}
    \begin{tabular}{@{}ll@{}}
        \toprule
        \textbf{Parameter} & \textbf{Value} \\
        \midrule
        LoRA rank ($r$)         & 8 \\
        LoRA alpha ($\alpha$)   & 16 \\
        LoRA dropout            & 0.0 \\
        Target modules          & All attention and MLP projections in \\ top 25\% of layers \\
        Gradient steps ($K$)    & 3 \\
        Optimizer               & AdamW~\citep{loshchilov2018decoupled} \\
        Learning rate ($\eta$)  & $10^{-4}$ \\
        Weight decay            & 0.0 \\
        Precision               & bfloat16 \\
        \bottomrule
    \end{tabular}
\end{table}

\subsection{Episodic Reset Protocol}
For query-level BSTTT, LoRA parameters are restored to
their initial values and the AdamW optimizer state is
cleared before each episode. After $K$ gradient steps
on the support set, the adapted model scores the held-out query by conditional log-likelihood over its answer
options. Parameter updates are not retained across episodes. For the trial-level DynToM variants, Scratchpad-TTT and Hierarchical-TTT, adaptation is shared within a trial and reset between trials.

\subsection{Action Reconstruction (AR) vs.\ Next-Token Loss (NTL)}

\paragraph{AR (Eq.~\ref{eq:ar_loss}).}
For each support example, we construct a multiple-choice
prompt containing the story, question, and candidate
answers, and compute each option's conditional
log-likelihood. The loss is the negative log-probability
of the correct option under a softmax over these scores,
averaged across support examples. This objective directly
supervises answer selection.

\paragraph{NTL (Eq.~\ref{eq:ntl_loss}).}
We apply standard causal language modeling to the
narrative text of the support examples, excluding
padding tokens from the loss. This condition provides
adaptation on episode narratives without answer supervision. Both objectives use the same LoRA configuration, learning rate, and number of gradient steps.
Their comparison evaluates answer-supervised adaptation
against narrative-only next-token adaptation.

\subsection{Scoring Protocol}

All methods use log-likelihood scoring for answer selection.
For each answer option $o_c$, we compute $\log p(o_c \mid \text{prompt}; \theta)$ by summing per-token log-probabilities.
The option with the highest total log-likelihood is selected.
This avoids parsing artifacts that can arise from free-form generation and ensures consistent evaluation across all methods.

\subsection{Compute Resources}
\label{app:compute}

Experiments used NVIDIA B200, A100, and H100 GPUs.
Table~\ref{tab:compute} summarizes the per-benchmark compute costs.

\begin{table}[t]
    \caption{\textbf{Evaluation time in minutes for one seed.}
    Frozen and BSTTT-AR values are estimates for an
    8B-class model on one B200 GPU, with $K{=}3$
    adaptation steps for BSTTT-AR. ICL values are recorded Llama-3.1-8B runtimes. GPU identifies the hardware used for ICL.}
    \label{tab:compute}
    \centering
    \small
    \begin{tabular}{@{}lrrrl@{}}
        \toprule
        \textbf{Benchmark} & \textbf{Frozen}
        & \textbf{BSTTT-AR} & \textbf{ICL}
        & \textbf{GPU} \\
        \midrule
        SimpleToM & 6 & 25 &  3.9 & A100 \\
        Hi-ToM    & 2 &  8 &  9.9 & H100 \\
        OpenToM   & 4 & 15 &  3.4 & H100 \\
        DynToM    & 6 & 22 & 54.5 & A100 \\
        \bottomrule
    \end{tabular}
\end{table}

\paragraph{Per-episode overhead.}
Each BSTTT episode involves: (1)~attaching and zero-initializing rank-$8$ LoRA modules to the top 25\% of transformer layers, (2)~$K{=}3$ forward--backward passes on the support set ($|\mathcal{S}| \in \{1, 4\}$ items), and (3)~one forward pass for query scoring.
For a 7B/8B model, a single episode takes approximately 0.4~seconds; for the 20B model, approximately 0.9~seconds.
The episodic reset (re-initializing LoRA weights and optimizer state) adds negligible overhead ($<$50\,ms per episode).

\paragraph{Preliminary experiments.}
The total compute for the research project, including hyperparameter exploration, pilot studies on smaller dataset subsets, and failed configurations, was approximately 4$\times$ the compute reported above (around 100+ GPU-hours total).

%% ============================================================
%% APPENDIX E: ADDITIONAL ABLATION STUDIES
%% ============================================================
\section{Ablation Studies}
\label{app:ablations}

The main text compares AR and NTL under the same
adaptation configuration (Table~\ref{tab:ar_vs_ntl}).
Here we report additional scaffold comparisons and
uncertainty estimates.

\subsection{Bootstrap Confidence Intervals}
\label{app:bootstrap}

Table~\ref{tab:ci_simpletom} reports SimpleToM accuracy
by question type, averaged over three seeds, alongside
95\% percentile bootstrap confidence intervals
computed using 2{,}000 resamples.

\begin{table*}[h]
    \caption{\textbf{SimpleToM per-question-type accuracy (\%) with 95\% bootstrap CIs} ($n{=}1{,}147$ items per question type, 2{,}000 resamples).}
    \label{tab:ci_simpletom}
    \centering
    \begin{tabular}{@{}llccc@{}}
        \toprule
        \textbf{Model} & \textbf{Method} & \textbf{Mental-State} & \textbf{Behavior} & \textbf{Judgment} \\
        \midrule
        \multirow{6}{*}{Qwen-7B}
            & Frozen              & 83.4\;{\scriptsize [81.3, 85.6]} & 50.7\;{\scriptsize [47.9, 53.6]} & 74.0\;{\scriptsize [71.5, 76.5]} \\
            & CoT                 & 84.7\;{\scriptsize [82.6, 86.8]} & 46.6\;{\scriptsize [43.8, 49.4]} & 74.4\;{\scriptsize [71.8, 76.8]} \\
            & MS-Reminder         & 83.6\;{\scriptsize [81.4, 85.8]} & 62.0\;{\scriptsize [59.4, 64.8]} & 85.4\;{\scriptsize [83.3, 87.4]} \\
            & Matched ICL
& 81.2\;{\scriptsize [78.9, 83.4]}
& 60.5\;{\scriptsize [57.6, 63.5]}
& 45.4\;{\scriptsize [42.5, 48.4]} \\
            & BSTTT-AR            & 83.0\;{\scriptsize [80.8, 85.2]} & 55.0\;{\scriptsize [52.1, 57.8]} & 74.5\;{\scriptsize [71.9, 77.0]} \\
            & MS-Rem.+BSTTT       & 94.7\;{\scriptsize [93.4, 95.9]} & 66.4\;{\scriptsize [63.8, 69.0]} & 86.4\;{\scriptsize [84.3, 88.3]} \\
        \midrule
        \multirow{6}{*}{Llama-8B}
            & Frozen              & 99.4\;{\scriptsize [99.0, 99.8]} & 28.9\;{\scriptsize [26.3, 31.6]} & 26.2\;{\scriptsize [23.5, 28.7]} \\
            & CoT                 & 100.0\;{\scriptsize [100.0, 100.0]} & 32.7\;{\scriptsize [30.0, 35.4]} & 38.3\;{\scriptsize [35.4, 41.2]} \\
            & MS-Reminder         & 99.2\;{\scriptsize [98.7, 99.7]} & 63.2\;{\scriptsize [60.4, 66.0]} & 34.0\;{\scriptsize [31.2, 36.5]} \\
            & Matched ICL
& 99.4\;{\scriptsize [99.0, 99.8]}
& 66.0\;{\scriptsize [63.4, 68.8]}
& 14.0\;{\scriptsize [12.0, 16.1]} \\
            & BSTTT-AR            & 99.2\;{\scriptsize [98.7, 99.7]} & 37.1\;{\scriptsize [34.4, 39.9]} & 26.4\;{\scriptsize [23.8, 28.9]} \\
            & MS-Rem.+BSTTT       & 99.8\;{\scriptsize [99.6, 100.0]} & 69.7\;{\scriptsize [67.3, 72.3]} & 33.7\;{\scriptsize [31.0, 36.3]} \\
        \midrule
        \multirow{6}{*}{GPT-oss-20B}
            & Frozen              & 100.0\;{\scriptsize [100.0, 100.0]} & 69.6\;{\scriptsize [67.0, 72.2]} & 3.1\;{\scriptsize [2.2, 4.2]} \\
            & CoT                 & 100.0\;{\scriptsize [100.0, 100.0]} & 71.8\;{\scriptsize [69.1, 74.5]} & 8.0\;{\scriptsize [6.5, 9.7]} \\
            & MS-Reminder         & 100.0\;{\scriptsize [100.0, 100.0]} & 74.8\;{\scriptsize [72.2, 77.3]} & 17.6\;{\scriptsize [15.5, 19.9]} \\
            & Matched ICL
& 100.0\;{\scriptsize [100.0, 100.0]}
& 73.4\;{\scriptsize [70.7, 75.9]}
& 26.8\;{\scriptsize [24.3, 29.4]} \\
            & BSTTT-AR            & 100.0\;{\scriptsize [100.0, 100.0]} & 71.4\;{\scriptsize [68.8, 74.1]} & 5.3\;{\scriptsize [4.0, 6.7]} \\
            & MS-Rem.+BSTTT       & 100.0\;{\scriptsize [100.0, 100.0]} & 77.3\;{\scriptsize [74.8, 79.7]} & 18.7\;{\scriptsize [16.5, 21.0]} \\
        \bottomrule
    \end{tabular}
\end{table*}

Table~\ref{tab:ci_external} reports aggregate accuracy
and bootstrap confidence intervals for Hi-ToM and
OpenToM.

\begin{table*}[t]
    \caption{\textbf{Hi-ToM and OpenToM accuracy (\%)
    with 95\% bootstrap confidence intervals.} Results for GPT-oss-20B, using 2,000 resamples. Hi-ToM contains 1,200 evaluation items and OpenToM contains 2,000.}
    \label{tab:ci_external}
    \centering
    \begin{tabular}{@{}llcc@{}}
        \toprule
        \textbf{Benchmark} & \textbf{Method}
        & \textbf{Accuracy (\%)} & \textbf{95\% CI} \\
        \midrule
        \multirow{8}{*}{Hi-ToM}
            & Frozen           & 32.6 & [30.0, 35.1] \\
            & CoT              & 32.7 & [30.0, 35.2] \\
            & Scratchpad       & 34.8 & [32.2, 37.4] \\
            & BSTTT-NTL        & 31.6 & [28.9, 34.1] \\
            & BSTTT-AR         & 47.3 & [46.3, 48.3] \\
            & Matched ICL      & 39.5 & [36.7, 42.2] \\
            & Scratchpad+BSTTT & 46.5 & [43.5, 49.2] \\
            & SimToM+BSTTT     & 41.2 & [37.0, 42.7] \\
        \midrule
        \multirow{8}{*}{OpenToM}
            & Frozen           & 35.6 & [33.6, 37.8] \\
            & CoT              & 31.5 & [29.6, 33.5] \\
            & Scratchpad       & 35.8 & [33.7, 37.9] \\
            & BSTTT-NTL        & 35.1 & [33.1, 37.2] \\
            & BSTTT-AR         & 36.0 & [34.0, 38.2] \\
            & Matched ICL      & 28.9 & [27.0, 30.9] \\
            & SimToM           & 34.4 & [32.3, 36.4] \\
            & SimToM+BSTTT     & 34.9 & [32.3, 36.5] \\
        \bottomrule
    \end{tabular}
\end{table*}

\paragraph{Interpretation.}
The intervals support substantial but incomplete recovery.
For Llama-8B, behavior accuracy increases from
28.9\% [26.3, 31.6] under frozen inference to
69.7\% [67.3, 72.3] with MS-Reminder+BSTTT.
The 40.8~pp improvement is large relative to the
estimated sampling uncertainty.

For GPT-oss-20B, judgment accuracy rises from
3.1\% [2.2, 4.2] to 18.7\% [16.5, 21.0] with
MS-Reminder+BSTTT and 26.8\% [24.3, 29.4] with
matched ICL. Despite these gains, accuracy remains
below the 50\% chance level, showing that measurable
recovery does not imply reliable evaluative reasoning.

\section{Illustrative Benchmark Example}
\label{app:examples}

The following SimpleToM example~\citep{gu2026simpletom}
illustrates the distinction between mental-state
recognition, behavior prediction, and judgment within
a shared narrative.

\begin{tcolorbox}[
    colback=gray!3, colframe=gray!40, boxrule=0.3pt,
    left=4pt, right=4pt, top=3pt, bottom=3pt
]
\textbf{Story:}
The can of Pringles has moldy chips in it.
Mary picks up the can in the supermarket and
walks to the cashier.\\[4pt]

\textbf{Mental-state recognition:}
Is Mary likely to be aware that
``The can of Pringles has moldy chips in it''?\\
(A) No \qquad (B) Yes\\[4pt]

\textbf{Behavior prediction:}
What will Mary likely do next?\\
(A) Pay for the chips \qquad
(B) Report the moldy chips\\[4pt]

\textbf{Judgment:}
Mary paid for the chips. Was that reasonable?\\
(A) Reasonable \qquad (B) Not reasonable
\end{tcolorbox}

The mental-state question assesses whether the model
recognizes Mary's unawareness. The behavior and judgment
questions assess whether it uses that information to
predict her action and evaluate its reasonableness.
A model that identifies Mary's unawareness but predicts
that she will report the mold demonstrates successful
recognition without corresponding application.

\noindent\textbf{OpenToM example.}

This example~\citep{xu-etal-2024-opentom} contrasts
attitude inference with reasoning about the consequences
of an object's relocation.

\begin{tcolorbox}[
    colback=gray!3, colframe=gray!40, boxrule=0.3pt,
    left=4pt, right=4pt, top=3pt, bottom=3pt
]
\textbf{Narrative (condensed):}
Diego dislikes scarves, while Amir likes them.
Both see a scarf in a basket on the patio.
Diego moves the scarf from the basket into a donation
bin, and Amir witnesses the move.\\[4pt]

\textbf{Attitude question:}
As Amir, what is your attitude towards Diego's action
of moving scarf from basket to a donation bin?\\
(A) Negative \quad (B) Neutral \quad (C) Positive\\[4pt]

\textbf{Multihop first-order question:}
From Amir's perspective, how does basket's fullness
change by the end of the story?\\
(A) Equally accessible \quad (B) Equally full\\
(C) Less accessible \quad (D) Less full\\
(E) More accessible \quad (F) More full
\end{tcolorbox}

The attitude question requires inferring Amir's response
to the action from his preference. The multihop question
requires using his knowledge of the scarf's relocation
to infer a consequence for the basket.
These illustrate our distinction between mental-state
identification and further inference from an agent's
perspective.

\noindent\textbf{DynToM example.}

This example~\citep{xiao-etal-2025-towards} contrasts
identifying a local belief with tracking its evolution
across a five-scenario trajectory. Answer options are
omitted for brevity.

\begin{tcolorbox}[
    colback=gray!3, colframe=gray!40, boxrule=0.3pt,
    left=4pt, right=4pt, top=3pt, bottom=3pt
]
\textbf{Context:}
Henry Lozano, a business analyst, and Charles Beyene,
a security guard, work at the same hotel.\\[4pt]

\textbf{Trajectory (condensed):}\\
\textbf{1.} Henry notices suspicious activity in the
hotel lobby and raises his concerns with Charles.\\
\textbf{2.} Charles suggests a misunderstanding;
Henry believes Charles is underestimating the danger.\\
\textbf{3.} Further investigation reveals that the
guests were participating in a scavenger hunt,
resolving Henry's initial suspicion.\\
\textbf{4.} A staff member reports possible internal
thefts, prompting Henry to investigate hotel staff.\\
\textbf{5.} Henry identifies a staff member manipulating
inventory records and confronts them.\\[4pt]

\textbf{Understanding question (\texttt{type\_a}):}
What is the belief of Henry Lozano in scenario 1?\\[4pt]

\textbf{Transformation question (\texttt{type\_d}):}
How does the belief of Henry Lozano change across
the 5 scenarios?
\end{tcolorbox}

The understanding question concerns a belief at one
point in the narrative. The transformation question
requires integrating successive revisions, including
the resolution of the initial suspicion and the
emergence of a different explanation.

%% ============================================================
%% APPENDIX G: CONSOLIDATED PER-MODEL RESULTS
%% ============================================================
\section{Consolidated Per-Model Results}
\label{app:consolidated_results}

Tables~\ref{tab:main_simpletom_datasetwise},
\ref{tab:main_hitom_datasetwise},
\ref{tab:main_opentom_datasetwise}, and
\ref{tab:main_dyntom_datasetwise} report per-model
results for SimpleToM, Hi-ToM, OpenToM, and DynToM,
respectively. They complement the cross-model summary
in Table~\ref{tab:summary} by reporting the individual
method configurations.

\subsection{Cross-Scenario In-Context Learning}
\label{app:cross_scenario_icl}

To examine dependence on scenario-specific support
information, we evaluate ICL with answered questions
from other scenario families or trials. Support
selection excludes the query's family or trial and
overlapping normalized narratives, while retaining
the original support counts: one for SimpleToM and
four for the other benchmarks. Model parameters
remain fixed.

\begin{table*}[t]
\caption{\textbf{Cross-scenario ICL results across models.}
Accuracy (\%) with answered supports drawn from other
scenario families or trials.}
\label{tab:cross_scenario_icl}
\centering
\begin{tabular*}{\textwidth}{@{\extracolsep{\fill}}lcccc@{}}
\toprule
\textbf{Model}
& \textbf{SimpleToM}
& \textbf{Hi-ToM}
& \textbf{OpenToM}
& \textbf{DynToM} \\
\midrule
Llama-3.1-8B & 57.37 & 42.50 & 48.50 & 30.41 \\
Qwen2.5-7B  & 59.46 & 48.17 & 42.26 & 39.20 \\
GPT-OSS-20B & 71.87 & 29.33 & 27.80 & 13.60 \\
\bottomrule
\end{tabular*}
\end{table*}

% \begin{table}[t]
% \centering
% \caption{\textbf{Cross-scenario ICL accuracy (\%).}
% Columns denote SimpleToM (S), Hi-ToM (H), OpenToM (O),
% and DynToM (D).}
% \label{tab:cross_scenario_icl}
% \begin{tabular}{@{}lrrrr@{}}
% \toprule
% \textbf{Model} & \textbf{S} & \textbf{H}
% & \textbf{O} & \textbf{D} \\
% \midrule
% Llama-3.1-8B & 57.37 & 42.50 & 48.50 & 30.41 \\
% Qwen2.5-7B   & 59.46 & 48.17 & 42.26 & 39.20 \\
% GPT-OSS-20B  & 71.87 & 29.33 & 27.80 & 13.60 \\
% \bottomrule
% \end{tabular}
% \end{table}

This control evaluates performance with demonstrations
that exclude the query's narrative.

For GPT-OSS on SimpleToM, cross-scenario ICL achieves
100.0\% mental-state accuracy, 73.4\% behavior accuracy, and 42.2\% judgment accuracy, averaging 71.87\% over 1,147 questions per type. Compared with the reported matched ICL results, the difference is
concentrated in judgment (26.8\% versus 42.2\%),
while recognition and behavior remain essentially
unchanged.

\paragraph{Method names.}
BSTTT-AR and BSTTT-NTL denote adaptation with vanilla
query prompting; composed methods explicitly name
the accompanying prompt intervention.
MS-Reminder denotes the SimpleToM mental-state
reminder, while Scratchpad-Frozen denotes DynToM's
table of predicted states.
Matched ICL uses answered demonstrations in context
without parameter updates.
Benchmark-specific implementations are described
in Appendix~\ref{app:method_details} and summarized
in Table~\ref{tab:appendix_method_map}.

\begin{table*}[t]
\caption{\textbf{SimpleToM results across models.}
Overall accuracy (\%) across 3,441 evaluation episodes,
with 1,147 per question type.
Bold identifies the best result for each model;
underlining identifies the best result without
parameter updates.
MS-Reminder+BSTTT-AR combines the mental-state
reminder with AR adaptation
(Appendix~\ref{app:simpletom_methods}).}
\label{tab:main_simpletom_datasetwise}
\centering
\begin{tabular}{@{}lccc@{}}
\toprule
\textbf{Method}
& \textbf{Qwen-2.5-7B}
& \textbf{Llama-3.1-8B}
& \textbf{GPT-oss-20B} \\
\midrule
Frozen                  & 69.4 & 51.5 & 57.6 \\
CoT                     & 68.5 & 57.0 & 59.9 \\
MS-Reminder             & \underline{77.0}
                        & \underline{65.5} & 64.1 \\
Matched ICL             & 62.4 & 59.8
                        & \underline{\textbf{66.7}} \\
BSTTT-NTL               & 69.0 & 51.7 & 56.9 \\
BSTTT-AR                & 70.8 & 54.2 & 58.9 \\
MS-Reminder+BSTTT-AR     & \textbf{82.5}
                        & \textbf{67.7} & 65.4 \\
\bottomrule
\end{tabular}
\end{table*}

\begin{table*}[t]
\caption{\textbf{Hi-ToM results across models.}
Accuracy (\%) on 1,200 evaluation items over three seeds. Bold identifies the best result for each model;
underlining identifies the best result without
parameter updates. Hi-ToM evaluates factual and recursive belief questions without a paired explicit--applied decomposition (Appendix~\ref{app:bench_hitom}).}
\label{tab:main_hitom_datasetwise}
\centering
\begin{tabular}{@{}lccc@{}}
\toprule
\textbf{Method}
& \textbf{Qwen-2.5-7B}
& \textbf{Llama-3.1-8B}
& \textbf{GPT-oss-20B} \\
\midrule
Frozen                  & 40.7 & 41.4 & 32.6 \\
CoT                     & 42.6 & 38.8 & 32.7 \\
Scratchpad              & 43.8 & 43.0 & 34.8 \\
SimToM                  & 38.8 & 41.8 & 31.7 \\
SymbolicToM             & 43.0 & 47.5 & 25.7 \\
Matched ICL             & \underline{\textbf{60.6}}
                        & \underline{\textbf{60.3}}
                        & \underline{39.5} \\
BSTTT-NTL               & 40.7 & 41.4 & 31.6 \\
BSTTT-AR                & 40.7 & 54.4 & \textbf{47.3} \\
Scratchpad+BSTTT-AR     & 53.7 & 53.7 & 46.5 \\
SimToM+BSTTT-AR         & 49.1 & 53.9 & 41.2 \\
SymbolicToM+BSTTT-AR    & 52.7 & 55.3 & 32.5 \\
\bottomrule
\end{tabular}
\end{table*}

\begin{table*}[t]
\caption{\textbf{OpenToM results across models.}
Accuracy (\%) on 2,000 evaluation items over three-seed averages. Bold identifies the best result for each model; underlining identifies the best result without parameter updates.}
\label{tab:main_opentom_datasetwise}
\centering
\begin{tabular}{@{}lccc@{}}
\toprule
\textbf{Method}
& \textbf{Qwen-2.5-7B}
& \textbf{Llama-3.1-8B}
& \textbf{GPT-oss-20B} \\
\midrule
Frozen                  & 40.5 & 42.5 & 35.6 \\
CoT                     & 41.0 & 41.6 & 31.5 \\
Scratchpad              & 40.4 & 45.8 & \underline{35.8} \\
SimToM                  & 45.6 & 48.1 & 34.4 \\
SymbolicToM             & 42.7 & 48.8 & 28.9 \\
Matched ICL             & \underline{\textbf{60.4}}
                        & \underline{\textbf{56.4}}
                        & 28.85 \\
BSTTT-NTL               & 40.5 & 42.7 & 35.1 \\
BSTTT-AR                & 40.5 & 43.1 & 36.0 \\
Scratchpad+BSTTT-AR     & 36.9 & 43.2 & \textbf{36.3} \\
SimToM+BSTTT-AR         & 46.7 & 47.7 & 34.9 \\
SymbolicToM+BSTTT-AR    & 39.2 & 50.0 & 27.5 \\
\bottomrule
\end{tabular}
\end{table*}

\begin{table*}[t]
\caption{\textbf{DynToM results across models.}
Accuracy (\%) on 3,000 evaluation items report three-seed averages. Bold identifies the best result for each model; underlining identifies the best result without parameter updates, excluding the oracle control. Scratchpad-Oracle supplies annotated mental-state tables and is reported separately.}
\label{tab:main_dyntom_datasetwise}
\centering
\begin{tabular}{@{}lccc@{}}
\toprule
\textbf{Method}
& \textbf{Qwen-2.5-7B}
& \textbf{Llama-3.1-8B}
& \textbf{GPT-oss-20B} \\
\midrule
Frozen              & 35.7 & 19.7 & 13.0 \\
CoT                 & 35.7 & 24.5
                    & \underline{\textbf{14.4}} \\
Scratchpad-Frozen   & 38.4 & 20.8 & 14.3 \\
Matched ICL         & \underline{\textbf{39.9}}
                    & \underline{\textbf{32.6}} & 13.3 \\
BSTTT-NTL           & 35.7 & 19.9 & 13.2 \\
BSTTT-AR            & 35.4 & 20.7 & 13.2 \\
Scratchpad-TTT      & 38.8 & 22.2 & 14.1 \\
Hierarchical-TTT    & 37.9 & 16.0 & 12.4 \\
\midrule
Scratchpad-Oracle   & 42.2 & 22.0 & 14.5 \\
\bottomrule
\end{tabular}
\end{table*}

\subsection{Cross-Model Patterns in Full Results}

The consolidated tables show that recovery depends
on both the model and the intervention.

\paragraph{Hi-ToM improves through both adaptation and ICL.}
The strongest BSTTT configuration improves accuracy
over frozen inference by 13.0--14.7~pp across models.
However, matched ICL achieves the highest accuracy
for Qwen and Llama, while BSTTT-AR remains strongest
for GPT-oss. Recursive belief reasoning therefore
shows substantial recoverability, but weight updates
are not uniformly the most effective intervention.

\paragraph{Prompt scaffold rankings vary across models.}
On Hi-ToM, SymbolicToM is the strongest structured
prompt for Llama (47.5\%), whereas Scratchpad performs
best for Qwen and GPT-oss. Composition also depends
on the model: Qwen benefits substantially from
Scratchpad+BSTTT, while GPT-oss performs better with
BSTTT-AR alone than with any tested prompting
composition. These differences motivate comparing
each composition against both constituent methods.

\paragraph{DynToM shows uneven temporal recovery.}
Matched ICL raises overall accuracy from 19.7\% to
32.6\% for Llama and from 35.7\% to 39.9\% for Qwen.
GPT-oss benefits little from the tested interventions.
The remaining errors, including those under
Scratchpad-Oracle, indicate that supplying local
state information is insufficient for reliable
performance. They do not establish an intrinsic
limit on temporal reasoning.

\paragraph{SimpleToM favors different interventions across models.}
MS-Reminder+BSTTT achieves the highest overall accuracy
for Qwen and Llama, whereas matched ICL is strongest
for GPT-oss. The composed intervention improves over
the reminder alone for all three models, showing
that weight adaptation can contribute beyond making
mental-state information explicit. Per-subtype results
are needed to determine which competencies account
for these aggregate gains.

\paragraph{OpenToM recovery does not require explicit scaffolds.}
Matched ICL outperforms every evaluated BSTTT
configuration for Qwen and Llama, reaching 60.4\%
and 56.4\%, respectively. GPT-oss shows the opposite
pattern: ICL underperforms frozen inference, while
BSTTT and its compositions provide small gains.
Thus, multihop recovery depends on how a model uses
the supplied information; it cannot be attributed
exclusively to structured prompting or weight updates.

\subsection{Per-Subtype and Gap Recovery Breakdown}

Table~\ref{tab:simpletom_subtypes} separates mental-state recognition, behavior prediction, and judgment.
For Llama, MS-Reminder+BSTTT-AR raises behavior accuracy from 28.9\% to 69.7\%, while mental-state accuracy remains at 99.4\%. Matched ICL also improves behavior prediction, reaching 66.0\%, but reduces judgment accuracy from 26.2\% to 14.0\%. Recovery in behavior prediction therefore does not necessarily transfer to evaluating the reasonableness of an action.

GPT-oss exhibits a different pattern: recognition
remains at ceiling, while judgment accuracy improves
from 3.1\% to 18.7\% with MS-Reminder+BSTTT-AR and
26.8\% with ICL. For Qwen, the composed intervention
improves all three question types, whereas ICL improves
behavior prediction but reduces recognition and
judgment accuracy. These differences show why overall
accuracy alone cannot establish which aspects of
ToM an intervention improves.

\begin{table*}[t]
\caption{\textbf{SimpleToM accuracy (\%) by question type.} Results are averaged over three seeds.
MS-Reminder+BSTTT-AR combines mental-state prompting
with parameter adaptation. The decomposition distinguishes improvements in behavior prediction from changes in recognition and judgment.}
\label{tab:simpletom_subtypes}
\centering
\begin{tabular}{@{}llccc@{}}
\toprule
\textbf{Model} & \textbf{Method}
& \textbf{Mental-state}
& \textbf{Behavior}
& \textbf{Judgment} \\
\midrule
\multirow{3}{*}{Qwen-2.5-7B}
& Frozen                  & 83.4 & 50.7 & 74.0 \\
& Matched ICL             & 81.2 & 60.5 & 45.4 \\
& MS-Reminder+BSTTT-AR     & 94.7 & 66.4 & 86.4 \\
\midrule
\multirow{3}{*}{Llama-3.1-8B}
& Frozen                  & 99.4 & 28.9 & 26.2 \\
& Matched ICL             & 99.4 & 66.0 & 14.0 \\
& MS-Reminder+BSTTT-AR     & 99.8 & 69.7 & 33.7 \\
\midrule
\multirow{3}{*}{GPT-oss-20B}
& Frozen                  & 100.0 & 69.6 & 3.1 \\
& Matched ICL             & 100.0 & 73.4 & 26.8 \\
& MS-Reminder+BSTTT-AR     & 100.0 & 77.3 & 18.7 \\
\bottomrule
\end{tabular}
\end{table*}

\begin{table*}[t]
\centering
\caption{\textbf{Values underlying
Figure~\ref{fig:gap_recovery}.}
Accuracy (\%) on the explicit and applied subsets.
Explicit accuracy is reported under frozen inference;
applied accuracy pools the applied question types.
``Best BSTTT'' denotes the BSTTT configuration selected
by overall benchmark accuracy. $\Delta_{\mathrm{gap}}$ is frozen explicit accuracy minus frozen applied accuracy, in percentage points. Hi-ToM reports overall accuracy because it lacks a paired explicit--applied decomposition. Intervention gains are obtained by subtracting the corresponding frozen accuracy.}
\label{tab:fig2_gap_recovery_values}

\begin{tabular}{@{}llrrrrrr@{}}
\toprule
& & \textbf{Explicit}
& \multicolumn{4}{c}{\textbf{Applied}}
& \\
\cmidrule(lr){3-3}
\cmidrule(lr){4-7}
\textbf{Benchmark} & \textbf{Model}
& \textbf{Frozen}
& \textbf{Frozen}
& \textbf{BSTTT-AR}
& \textbf{ICL}
& \textbf{Best BSTTT}
& $\boldsymbol{\Delta_{\mathrm{gap}}}$ \\
\midrule
\multirow{3}{*}{SimpleToM}
& Qwen-2.5-7B  & 83.4  & 62.4 & 64.7 & 53.0 & 76.4 & 21.0 \\
& Llama-3.1-8B & 99.4  & 27.5 & 31.8 & 40.0 & 51.7 & 71.9 \\
& GPT-oss-20B  & 100.0 & 36.4 & 38.4 & 50.1 & 48.0 & 63.6 \\
\midrule
\multirow{3}{*}{OpenToM}
& Qwen-2.5-7B  & 46.5 & 38.0 & 35.7 & 66.5 & 47.4 & 8.5 \\
& Llama-3.1-8B & 46.1 & 40.9 & 41.0 & 61.5 & 51.8 & 5.2 \\
& GPT-oss-20B  & 35.1 & 35.9 & 36.3 & 26.2 & 36.7 & $-0.8$ \\
\midrule
\multirow{3}{*}{Hi-ToM}
& Qwen-2.5-7B  & --- & 40.7 & 40.7 & 60.6 & 53.7 & --- \\
& Llama-3.1-8B & --- & 41.4 & 54.4 & 60.3 & 55.3 & --- \\
& GPT-oss-20B  & --- & 32.6 & 47.3 & 39.5 & 47.3 & --- \\
\midrule
\multirow{3}{*}{DynToM}
& Qwen-2.5-7B  & 30.5 & 37.8 & 37.7 & 38.5 & 41.5 & $-7.3$ \\
& Llama-3.1-8B & 29.6 & 15.9 & 17.4 & 26.8 & 17.7 & 13.7 \\
& GPT-oss-20B  & 16.6 & 11.7 & 12.5 & 12.1 & 13.5 & 4.9 \\
\bottomrule
\end{tabular}
\end{table*}

\end{document}